\documentclass[10pt,journal,compsoc]{IEEEtran}
\makeatletter
\def\ps@IEEEtitlepagestyle{%
  \def\@oddfoot{\mycopyrightnotice}%
  \def\@oddhead{\hbox{}\@IEEEheaderstyle\leftmark\hfil\thepage}\relax
  \def\@evenhead{\@IEEEheaderstyle\thepage\hfil\leftmark\hbox{}}\relax
  \def\@evenfoot{}%
}
\def\mycopyrightnotice{%
  \begin{minipage}{\textwidth}
  \centering \scriptsize
  Copyright~\copyright~20XX IEEE.  Personal use of this material is permitted.  Permission from IEEE must be obtained for all other uses, in any current or future media, including reprinting/republishing this material for advertising or promotional purposes, creating new collective works, for resale or redistribution to servers or lists, or reuse of any copyrighted component of this work in other works.
  \end{minipage}
}
\makeatother

\ifCLASSOPTIONcompsoc
  \usepackage[nocompress]{cite}
\else
  \usepackage{cite}
\fi
\ifCLASSINFOpdf
\else
\fi

\usepackage{url}
\usepackage{amsmath}
\usepackage{graphicx}  
\usepackage{subfigure}
\usepackage{booktabs}
\usepackage{ragged2e}
\usepackage{multirow}
\usepackage{colortbl}
\usepackage{xcolor}
\usepackage{amsmath,amssymb,bbm}
\usepackage{lipsum}
\usepackage{pgfplots}
\usepackage{algorithm}
\usepackage{algpseudocode}
\usepackage{enumitem}
\usepackage{bm}
\usepackage{soul}
\usepackage{longtable}
\usepackage{arydshln}
\usepackage{makecell}
\usepackage[numbers]{natbib}
\usepackage[encapsulated]{CJK}
\usepackage{booktabs}
\usepackage[flushleft]{threeparttable}

\begin{document}
\title{Universal Multimodal Representation for Language Understanding}

\begin{CJK*}{UTF8}{gkai}

\author{Zhuosheng Zhang$^{\#}$, Kehai Chen, Rui Wang$^{\#}$, Masao Utiyama, Eiichiro Sumita, Zuchao Li, Hai Zhao*
	\IEEEcompsocitemizethanks{
	\IEEEcompsocthanksitem{Z. Zhang, R. Wang, Z. Li, H. Zhao are with the Department of Computer Science and Engineering, Shanghai Jiao Tong University, China and also with Key Laboratory of Shanghai Education Commission for Intelligent Interaction and Cognitive Engineering, Shanghai Jiao Tong University, China. K. Chen is with the School of Computer Science and Technology, Harbin Institute of Technology, Shenzhen, China. M. Utiyama and E. Sumita are with the National Institute of Information and Communications Technology (NICT), Japan. E-mail: 
		\{zhangzs, charlee\}@sjtu.edu.cn;
		chenkehai@hit.edu.cn;
		\{mutiyama,  eiichiro.sumita\}@nict.go.jp; wangrui.nlp@gmail.com; zhaohai@cs.sjtu.edu.cn. }
		\IEEEcompsocthanksitem{H. Zhao is supported by Key Projects of National Natural Science Foundation of China (U1836222 and 61733011). R. Wang is supported by National Natural Science Foundation of China (No. 6217020129), Shanghai Pujiang Program (No. 21PJ1406800), Shanghai Municipal Science and Technology Major Project (No. 2021SHZDZX0102), Beijing Academy of Artificial Intelligence (BAAI) (No. 4), CCF-Baidu Open Fund (No. CCF-BAIDU OF2022018). K. Chen is supported by National Natural Science Foundation of China (No. 62276077) and Shenzhen College Stability Support Plan (No. GXWD20220811170358002 and GXWD20220817123150002). }
		\IEEEcompsocthanksitem{Z. Zhang and R. Wang contribute equally to this work. Part of this work was finished when Z. Zhang and Z. Li visited NICT, and R. Wang was with NICT. Corresponding author: Hai Zhao.}
		}}

\IEEEtitleabstractindextext{%
\begin{abstract}
\justifying 
Representation learning is the foundation of natural language processing (NLP). This work presents new methods to employ visual information as assistant signals to general NLP tasks. For each sentence, we first retrieve a flexible number of images either from a light topic-image lookup table extracted over the existing sentence-image pairs or a shared cross-modal embedding space that is pre-trained on out-of-shelf text-image pairs. Then, the text and images are encoded by a Transformer encoder and convolutional neural network, respectively. The two sequences of representations are further fused by an attention layer for the interaction of the two modalities. In this study, the retrieval process is controllable and flexible. The universal visual representation overcomes the lack of large-scale bilingual sentence-image pairs. Our method can be easily applied to text-only tasks without manually annotated multimodal parallel corpora. We apply the proposed method to a wide range of natural language generation and understanding tasks, including neural machine translation, natural language inference, and semantic similarity. Experimental results show that our method is generally effective for different tasks and languages. Analysis indicates that the visual signals enrich textual representations of content words, {provide fine-grained grounding information about the relationship between concepts and events}, and potentially conduce to disambiguation.

\end{abstract}

}

\maketitle

\IEEEdisplaynontitleabstractindextext

\IEEEpeerreviewmaketitle

\IEEEraisesectionheading{\section{Introduction}\label{sec:introduction}}
\IEEEPARstart{L}{earning} contextualized representations of human languages is one of the major themes in natural language processing (NLP), which is also fundamental to training machines to understand human languages and handle advanced tasks, such as machine translation, question answering, and human-computer conversations. Text representation learning has evolved from standard distributed representations \cite{mikolov2013distributed,pennington2014glove} to contextualized language representation from deep pre-trained representation models (PRMs) \cite{Peters2018ELMO,radford2018improving,devlin2018bert,yang2019xlnet}. Despite the success of PRMs, NLP models commonly model the world knowledge (e.g, commonsense, rules, events, assertions extracted from raw texts) solely from textual features without grounding of the outside world, such as visual conception \cite{zhang2018image}. 
Languages are abstract and rather difficult for the brain to retain, whereas visuals are concrete and, as such, more easily remembered \cite{mcdaniel1986bizarre,meier2000accelerated}. Adopting multimodality would be essential for better background perception. Therefore, a trend of research has been motivated to apply non-linguistic modalities to language representations \cite{zhang2018image,ive2019distilling,shi2019visually,8269806,8691415,9215037}. 
	
	Most of previous works focus on joint modeling images and texts, involving vision-language (VL) pre-training \cite{su2019vl,lu2019vilbert,tan2019lxmert,li2019unicoder,zhou2019unified,sun2019videobert} and multimodal (MM) application tasks \cite{zablocki2018learning,wu2019glyce,9716741,8986691,9215037}. 
	However, these studies rely on large-scale text-image annotations as the paired input and thus are confined to VL or MM tasks, such as image captioning and visual question answering. 
	It is natural to boost the performance on VL and MM tasks as the concerned datasets are human-labeled with high quality. 
	However, the essential challenge lies with the real-world scenario as there is no such high-quality annotated text-image aligned corpus for text-only NLP applications. 
    Therefore, it is critical to investigate a general method to take advantage of visual information in a wide range of mono-modal (e.g., text-only) tasks. In addition, it is still not clear the role of images in language representation, as well as how to apply the multimodality in the standard NLP scenario.
	
	Taking multimodal machine translation (MMT) as an example, the starting point is to leverage visual information to improve the quality of the translation from the source to the target languages. However, the effectiveness heavily relies on the availability of bilingual parallel sentence pairs with manual image annotations, which hinders the image applicability to neural machine translation (NMT). As a result, the visual information is only applied to the translation task over specific multimodal datasets \cite{elliott2016multi30k,li2019coco,yoshikawa2017stair,miyazaki2016cross,hewitt2018learning}, instead of general text-only NMT~\cite{bahdanau2014neural,gehring-etal-2017-convolutional,NIPS2017_7181} and low-resource text-only NMT~\cite{fadaee2017data,lample2018phrase,ma2019flowseq,zhou2019handling}. In addition, because of the high cost of annotation, the content of one bilingual parallel sentence pair is paired with a single image, which is weak in capturing the diversity of visual information. Therefore, the current study of introducing visual information falls into a bottleneck in the multimodal NMT and is not feasible for text-only NMT and low-resource NMT.
	
	Our previous work \cite{Zhang2020Neural} finds that using monolingual corpora with image annotations can overcome the lack of large-scale bilingual sentence-image pairs, thereby extending image applicability in NMT, with performance gains. The method of using a lookup table is task-agnostic. This work stimulates our further thinking, and we are interested in answering the three major aspects of questions:
	
	(i) \textbf{Global Multimodality}: Can we apply the multimodality to standard text-only NLP tasks to enhance the language representations (Section \ref{sec:result}), e.g., natural language generation (Section \ref{sec:nmt}) and natural language understanding (Section \ref{sec:nlu})?
	
	(ii) \textbf{Interpretability}: Why does the universal representation method work (Section \ref{sec:benefits})?  How does multimodality improve language representation, and what is the network learned(Section 6.2-6.6)?

    (iii) \textbf{Quality}: How to control the quality of visual-text alignment to reduce noise (Section \ref{sec:thred})?

	In this paper, we present a universal visual representation (UVR) method relying only on a seed set of task-independent annotations, instead of the existing approach that depends on large-scale task-specific image-text annotation, thus breaking the bottleneck of using visual information in standard text-only NLP tasks. For each sentence, we retrieve diverse images from either a light topic-image lookup table or pre-trained shared text-visual embedding space that is pre-trained on a large-scale of text-image pairs, to connect both the mono-modal paths of text and image embeddings. The text and images are encoded by Transformer language model (LM) and a pre-trained convolutional neural network (CNN), respectively. A simple and effective attention layer is then designed to fuse the two sequences of representations. 
	
	
	
	Our approach can be easily applied to text-only tasks without manually annotated multimodal parallel corpora. Therefore, our method is universal in terms of the task requirements, in contrast to the recent vision-language models that require large-scale and expensive annotation datasets for each downstream task. The proposed method is evaluated on 14 NLP benchmark datasets involving natural language inference (NLI), semantic similarity, text classification, and machine translation. The experiments and analysis verify the effectiveness of the proposed method. To summarize, our contributions are primarily three-fold:
	
	(i) This work studies the universal visual representation for language representation in a broader view of the natural language processing scenario. Besides neural machine translation, this work leverages visual information as assistant signals for general NLP tasks, with the focus on investigating the global multimodality for general NLP, interpretability of effectiveness, and quality control of using universal visual representation.
    
    (ii) For the technical side, this work proposes new methods of semantic sentence-image matching from a shared cross-modal space to give more accurately paired images as topic information. We also present a new multimodal representation framework and systematically study the two main instances, including the model with the original TF-IDF topic-image lookup table and the newly proposed one from the retrieval from cross-modal retrieval.
    
    (iii) Experiments are extended to 14 representative NLP tasks, which show the effectiveness of the proposed method. {A series of in-depth analyses indicate that the visual signals enrich textual representations of content words, provide fine-grained grounding information about the relationship between concepts and events, and potentially conduce to disambiguation.}
    
\section{Background}
\subsection{Vision-Language Integration}
This study is related to that of VL methods (VLMs).
	Recently, there has been a great deal of interest in integrating image presentations in pre-trained Transformer architectures \cite{su2019vl,lu2019vilbert,tan2019lxmert,li2019unicoder,zhou2019unified,sun2019videobert,li2020oscar}. The common strategy is to take a Transformer model \cite{NIPS2017_7181}, such as BERT, as the backbone and learn aligned representations of visual and language in a pre-training manner inspired by the masked language modeling mechanism in pre-trained language models \cite{devlin2018bert}. These studies require the annotation of task-dependent sentence-image pairs, which are limited to VL tasks, such as image captioning and visual question answering. 
	
	Two studies \cite{wu2019glyce,Zhang2020Neural} are closely related to general image-enhanced LM. Glyce \cite{wu2019glyce} proposes incorporating glyph vectors for Chinese character representations. However, it can only be used for Chinese and only involves single image enhancement. Regarding the technical part, previous methods only benefit from one image per sentence. We propose taking advantage of a group of similar images using a filtering mechanism to form a more fine-grained visual-aware context.
	Our early version of this work \cite{Zhang2020Neural} proposes using multiple images for NMT, based on a text-image lookup table trained over a sentence-image pair corpus. However, the number of images is fixed because of the lack of similarity measurement in the simple lookup method, which possibly makes the resulting model suffer from the noise of irrelevant images. This work is improved from the perspectives of motivation and technique. It is motivated by cross-modal semantic retrieval in the shared embedding space. It adopts a neural matching method with a similarity threshold to control the expected matching degree flexibly, which is generally applicable to a broader range of NLP tasks. In addition, we conduct an in-depth analysis to investigate how the visual modality helps text representation.
    
	\subsection{Visual-Semantic Embeddings}
	Another research line is language grounding for images whose major topic is multimodality and cross-modality between images and text. The major focus is to bridge the gap between text and images through building visual-semantic embeddings \cite{frome2013devise,karpathy2015deep,ren2016joint,7534740,9633153}. 
	
	Prior studies have verified that representations of images and text can be jointly leveraged to build visual-semantic embeddings in a shared representation space \cite{frome2013devise,karpathy2015deep,ren2016joint,mukherjee2016gaussian}. To this end, a popular approach is to connect both the mono-modal text and image encoding paths using fully connected layers \cite{wang2018learning,engilberge2018finding}. The shared deep embedding can be used for cross-modal retrieval; thus, it can associate sentence text with associated images. Partly inspired by this line of research, we are motivated to incorporate visual awareness into sentence modeling by retrieving a group of images for a given sentence.
	
	One of the first techniques to align two views of heterogeneous data is the canonical correlation analysis method \cite{hotelling1992relations}, in which linear projections defined on both sides are optimized to maximize the cross-correlation. Recent studies have followed the two-path architecture \cite{wang2018learning,engilberge2018finding}, in which the encoder consists of a joint embedding of textual and image representations extracted from both the images and corresponding caption. Notably, \citet{engilberge2018finding} adopts RNN to encode sentence embeddings in the same space with extracted image representations from CNN. \citet{portaz2019image} enhances cross-modal retrieval using multilingual text. Inspired by the previous success of visual-semantic embeddings, we apply neural image retrieval from the joint space to fetch a group of associated images.

\begin{figure*}
    	\centering
    	\includegraphics[width=0.92\textwidth]{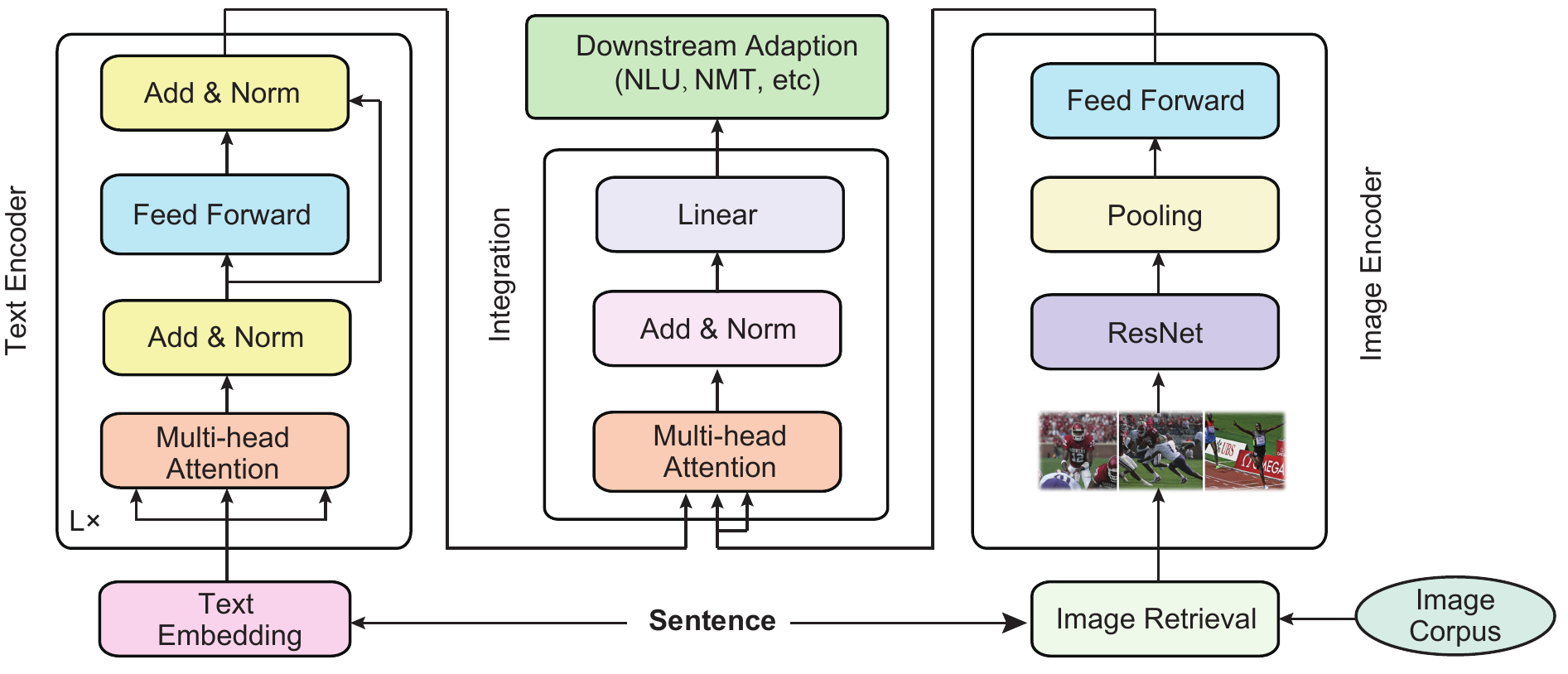}
    	\caption{Overview of the universal representation framework. Given a sentence as input, a group of related images will be retrieved by our image retrieval methods. The text and images are encoded by the text feature extractor and image feature extractor, respectively. Then the two sequences of representations are integrated using multi-head attention to form a joint representation in the same shape as the original text sequence representation. Finally, the joint representation is passed to downstream task-specific layers to give predictions.}
    	\label{overview-cmrm}
    \end{figure*}
    
	\section{Universal Representation Framework}
	This section overviews our universal representation framework. Given a sentence, we first fetch a group of matched images from our retrieval methods (details of our retrieval methods will be given in the next section). The text and images are encoded, respectively, by the text feature extractor and image feature extractor. Then the two sequences of representations are integrated using multi-head attention to form a joint representation, which is passed to downstream task-specific layers. Figure \ref{overview-cmrm} overviews the whole multimodal representation model. 
		
	\subsection{Encoding Layer}
	\subsubsection{Text Encoder}
	
	We pair each sentence with the top matched $m$ images according to the retrieval method above.
	Following \citep{devlin2018bert}, the sentence is fed into the multi-layer Transformer encoder \cite{NIPS2017_7181} to learn the text representation ${H} \in \mathbb{R}^{n \times d}$ where $n$ and $d$ are the input text length and dimension of hidden states for the text representation.
	
    {Let $X = \{x_1, \dots, x_n\}$ be the input sentence in length $n$.} We feed the sequence to a PRM encoder (e.g., BERT \citep{devlin2018bert}). In the encoder, the input sequence is firstly mapped to embeddings. Then, the embeddings are passed to multi-head attention layers \cite{NIPS2017_7181} to obtain the contextualized  representations, which is defined as
	\begin{align}\label{eq:mutihead}
	{H} = \textup{FFN}(\textup{MultiHead}(K,Q,V)),
	\end{align}
	where $K$, $Q$, $V$ are packed from the input sequence representation $X$. As the common practice, we set $K=Q=V$ in the implementation. \textup{MultiHead} is short for multi-head attention.
	
\subsubsection{Image Encoder}\label{sec:image_encoder}
Similar to the standard way of retrieving word embeddings, the image embeddings are fetched from a lookup table $\in R^{n_m \times d_m}$ that contains the image features encoded by a pre-trained ResNet \cite{he2016deep},\footnote{Note that this is the standard lookup table in embedding implementations, which is not our topic-image lookup table.} where $n_m$ is the number of the total number of unique images $+1$ and $d_m$ is the dimension of the image features.\footnote{We used the maxpooling layer of ResNet, which is in the size of $R^{n_m\times 2400}$.} The first row of the lookup table is filled by all-zero vectors, which will be used when no image is paired for the sentence.

After the feature lookup process, we obtain the image embeddings $E \in \mathbb{R}^{m \times d_m}$ for the $m$ input images. Then, the embeddings are passed to a feedforward layer, to produce the image representation ${M} \in \mathbb{R}^{m \times d}$ with the same hidden dimension as ${H}$:
\begin{equation} {{M}}=\textup{FFN}(\textup{ResNet}({{E}})).
\label{eq:resnet}
\end{equation}

There may exist cases when no word in the sentence can be found in the topic-image lookup table. When there is no paired image retrieved, we use the first-row all-zero vectors of the image lookup table as the ``blank features" in the intuition to tell the model to ignore them.

\subsection{Multimodal Integration Layer}

{We connect the visual and text modalities by calculating the attention between image and text features:
\begin{align}
    \alpha &= \textup{softmax}({H}({W}_g{M}+{b}_{g})^\top), \label{eq2:Image_Representation}\\
{H}' &= {\alpha}{M},\label{eq3}
\end{align}
where ${W}_g$ and ${b}_{g}$ are parameters to learn. {$\alpha \in \mathbb{R}^{n \times m}$ denotes the weights assigned to the different hidden states in the sentence and the image sequences.} ${H}'\in \mathbb{R}^{n \times d}$ is the weighted sum of all the hidden states and it represents how the sentence can be aligned to each hidden state in the image representation.}

The retrieval process may possibly introduce noise of irrelevant images. To alleviate the influence, we use a neural gating mechanism for information filtering. {In detail, we compute $\lambda \in [0, 1]^{n \times d}$} to weight the expected importance of image representation for each source word:
\begin{equation} \lambda=\textup{sigmoid}({W}_{\lambda}{{H}}'+{U}_{\lambda}{H}),
\label{eq3:Lamda_Image}
\end{equation}
where ${W}_{\lambda}$ and ${U}_{\lambda}$ are model parameters.
We then fuse ${H}$ and ${{H}}'$ and pass the resulting representation to layer normalization and learn an effective source representation:
\begin{equation} \hat{{H}}=\textup{LayerNorm}({H}+\lambda{{H}}').
\label{eq4:Fusing_Sourece_Representation}
\end{equation}

The text and image representations are jointly encoded as $\hat{{H}}$, which is fed to the task-specific layers for downstream decoding or predictions depending on task settings.

\subsection{Task-specific Layer}
In this section, we show how the joint representation $\hat{{H}}$ is used for downstream tasks by considering NMT and NLU tasks as examples, generally following the standard procedure of the concerned tasks. For NMT, $\hat{{H}}$ is directly fed to the decoder to learn a dependent-time context vector to predict the target translation. For other tasks, $\hat{{H}}$ is directly fed to a feed-forward layer to make the prediction, which follows the same downstream procedure as the Transformer-based LMs, like BERT \cite{devlin2018bert} and RoBERTa \cite{liu2019roberta}. Specifically, for sentence-pair tasks, we maintain the pairwise input as that in LMs and separate the encoded text representation into two individual sentence representations $\{{H}^1, {H}^2\}$, according to the positions. The two text representations are integrated with the corresponding image representations respectively $\{{M}^1, {M}^2\}$, and then the resulting sequences are concatenated  for prediction, $\hat{{H}} = \hat{{H}}^1 \circ \hat{{H}}^2$.

\section{Image Retrieval Methods}
	
	In this section, we describe our two visual retrieval models used for image retrieval given sentence text: 
	
	(i) \textbf{UVR-TILT}: retrieval by topic-image lookup table; 
	
    (ii) \textbf{UVR-CMRM}: retrieval from cross-modal embedding.

	\begin{figure*}
    	\centering
    	\includegraphics[width=1.0\textwidth]{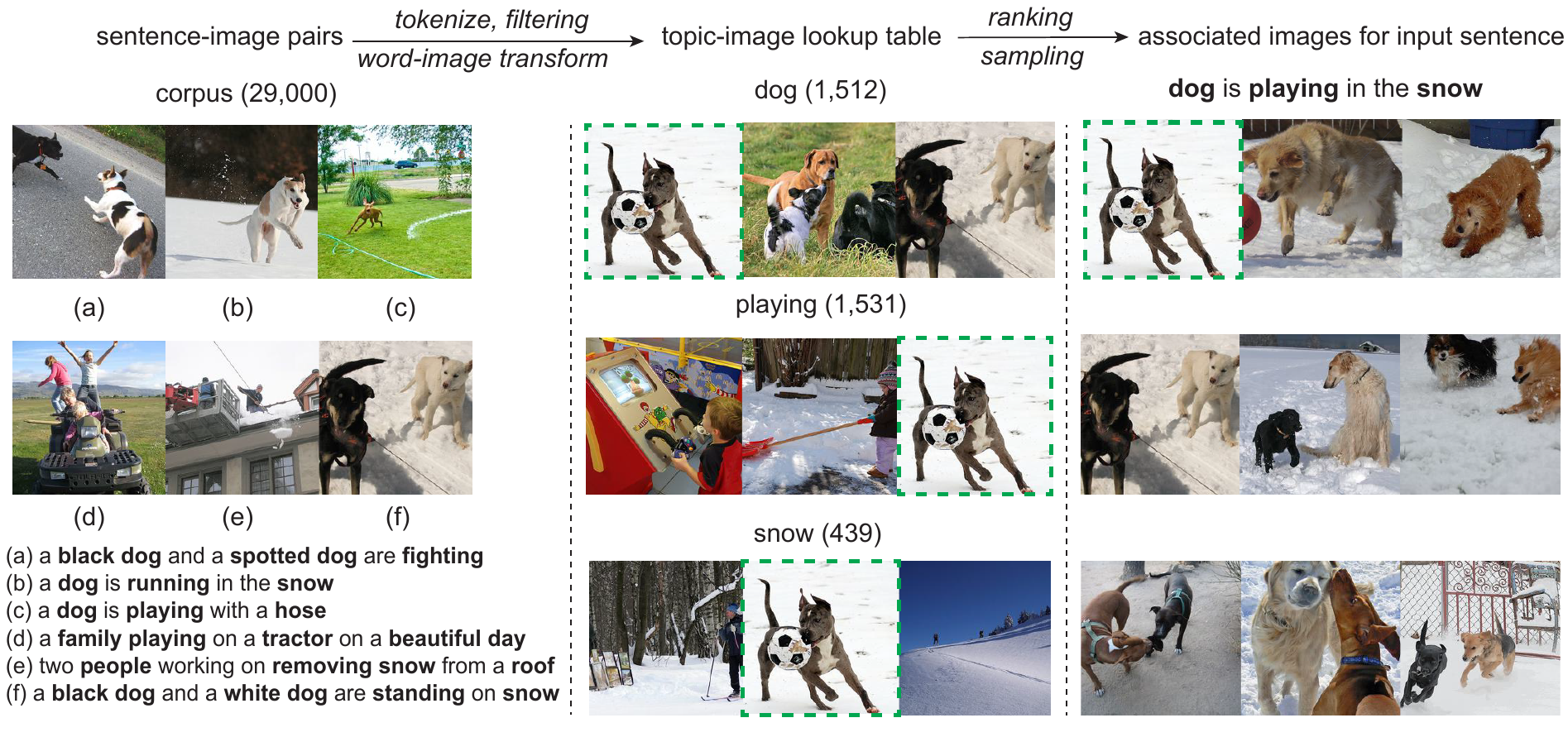}
    	\caption{\label{fig:img-sampling} Illustration of the TILT method. We first transform the existing sentence-image pairs from seed small-scale sentence-image datasets into a topic-image lookup table. For a given sentence, we extract its topic words and the associated images will be retrieved from the lookup table.}
    \end{figure*}
    
    \subsection{Model-I: Retrieval by Topic-Image Lookup Table}\label{fuzzy_method}
	
      \begin{algorithm}\small
    	\caption{Topic-image Lookup Table Conversion}\label{alg}
    	\begin{algorithmic}[1]
    		\Require Input sentences, $S=\{X_1, X_2, \dots\, X_I \}$ and paired images $E = \{e_1, e_2, \dots, e_I\}$
    		\Ensure Topic-image lookup table $\mathcal{Q}$ where each word is associated with a group of images
    		\State Obtain the TF-IDF dictionary $\mathcal{F}$ = TF-IDF($S$)
    		\State Transform sentence-image pair to topic-image lookup table $\mathcal{Q}$ = LookUp($S$, $E$, $\mathcal{F}$)
    		
    		\Procedure{TF-IDF}{$S$}
    		\For{each sentence in $S$}
    		\State Filter stop-words in the sentence
    		\State Calculate the TF-IDF weight for each word
    		\EndFor
    		\State \textbf{return} TF-IDF dictionary $\mathcal{F}$
    		\EndProcedure%
    		
    		\Procedure{LookUp}{$S$, $E$, $\mathcal{F}$}
    		\For{For each pair $\{X_i, e_i\} \in \texttt{zip}\{S, E\}$}
    		\State  Rank and pick out the top-$w$ ``topic" words in the sentence according to the TF-IDF score in the dictionary $\mathcal{F}$, and each sentence is reformed as $T=\{t_1, t_2, \dots, t_w\}$
    		\For{For each word $t_j$in $T$}
    		\If{$e_i$ not in $\mathcal{Q}[t_j]$}
    		\State Add $e_j$ to the corresponding image set $\mathcal{Q}[t_j]$ for word $t_j$
    		\EndIf
    		\EndFor
    		\EndFor
    		\State \textbf{return} Topic-image lookup table $\mathcal{Q}$ 
    		\EndProcedure%
    	\end{algorithmic}
    \end{algorithm}
    
    \subsubsection{Topic-image Lookup Table Conversion}
    In this section, we will introduce the proposed universal visual representation method.
    Our basic intuition is to transform the existing sentence-image pairs into a topic-image lookup table,\footnote{We use the training set of the \textit{Multi30K} dataset to build the topic-image lookup table.} which assumes the topic words in a sentence should be relevant to the paired image. The procedure can be seen as the inverted index where a topic word is mapped to a list of images.
    Consequently, a sentence can possess a group of images by retrieving the topic-image lookup table. 
    
    To focus on the major part of the sentence and suppress the noise such as stopwords and low-frequency words, we design a filtering method to extract the ``topic" words of the sentence through the term frequency-inverse document frequency (TF-IDF),\footnote{We describe our methods by regarding the processing unit as word though this method can also be applied to a subword-based sentence for which the subword is considered to be the processing unit.} inspired by \cite{chen2019neural}.
    Specifically, given an original input sentence $X=\{x_1, x_2, \dots, x_I\}$ of length $I$ and its paired image $e$, $X$ is first filtered by a stopword list,\footnote{\url{https://github.com/stopwords-iso/stopwords-en}.} and then the sentence is treated as a document $g$. We then compute TF-IDF $TI_{i,j}$ for each word $x_i$ in $g$,
    \begin{equation}
    TI_{i,j}=\frac{o_{i,j}}{\sum_{k}o_{k,j}}\times\textup{log}\frac{|G|}{1+|j:x_i\in g|},
    \label{eq23:tf-idf}
    \end{equation}
    where $o_{i,j}$ represents the number of occurrences of the word $x_i$ in the input sentence $g$, $|G|$ the total number of source language sentences in the training data, and $|j:x_i\in g|$ the number of source sentences including word $x_i$ in the training data.
    We then select the top-$w$ high TF-IDF words as the new image description $T=\{t_1, t_2, \dots, t_w\}$ for the input sentence.
    After the preprocessing, each filtered sentence $T$ is paired with an image $e$, and each word $t_i \in T$ is regarded as the topic word for image $e$. After processing the whole corpus (i.e.,  \textit{Multi30K}), we form a topic-image lookup table $\mathcal{Q}$ as described in Algorithm \ref{alg}, in which each topic word $t_i$ would be paired with dozens of images.
    

	
    \subsubsection{Image Retrieval}
    For the input sentence, we first obtain its topic words according to the text preprocessing method described above. Then we retrieve the associated images for each topic word from the lookup table $\mathcal{Q}$ and group all the retrieved images together to form an image list $\mathcal{G}$. We observe that an image might be associated with multiple topic words so that it would occur multiple times in the list $\mathcal{G}$. Thus, we sort the images according to the frequency of occurrences in $\mathcal{G}$ to maintain the same total number of images for each sentence at $m$. 
	
    Figure \ref{fig:img-sampling} illustrates the retrieval process.
    In the left block, we show six examples of sentence-image pairs in which the topic words are in boldface. 
    Then we process the corpus using the topic-image transformation method demonstrated above and obtain the topic-image lookup table. 
    For example, the word $dog$ is associated with 1,512 images. 
    For an input source sentence, we obtain the topic words (in boldface) using the same preprocessing. 
    Then we retrieve the corresponding images from the lookup table for each topic word. 
    Now we have a list of images, and some images appear multiple times as they have various topics (like the boxed image in Figure \ref{fig:img-sampling}). 
    So we sort the retrieved image list by the count of occurrence to pick out the top-$m$ images that cover the most topics of the sentence.

    At test time, the process of getting images is done using the image lookup table built by the training set, so we do not need to use the images from the validation and test sets in \textit{Multi30K} dataset.\footnote{The lookup table can be easily adapted to a wide range of other NLP tasks even without any paired image, and therefore opens our proposed model to generalization.}
    Intuitively, we do not strictly require the manual alignment of the word (or concept) and image but rely on the co-occurrence of the topic word and image, which is simpler and more general. 
    In this way, we call our method universal visual retrieval.

	\subsection{Model-II: Retrieval from Cross-modal Embedding}
	Following \citet{engilberge2018finding}, we train a semantic-visual embedding on a text-image corpus, which is then used for image retrieval. The semantic-visual embedding architecture comprises two paths to encode the text and images into vectors. Based on our preliminary experiments, we maintain the same settings in \citet{engilberge2018finding} by using the simple recurrent unit as text encoder, and the fully convolutional residual ResNet-152 \cite{xie2017aggregated} with Weldon pooling  \cite{durand2016weldon} as image encoder for our cross-modal retrieval model. 

	During training, each text $X$ is paired with (i) a positive image $Y$ that is paired with the text and (ii) a hard negative $Z$, which is selected as the image that has the highest similarity to the text while not being associated with it. Triplet loss \cite{wang2014learning,schroff2015facenet,gordo2017end} is used to enable the images to converge correctly to improve the performance of the proposed method: 
	\begin{equation}
	\label{eq:cost}
	loss(X, Y, Z) = max(0, \gamma - E(X) \cdot E(Y) + E(X) \cdot E(Z)),
	\end{equation}
	{where $E(X)$, $E(Y)$, and $E(Z)$ are the embeddings of $X$, $Y$, and $Z$, respectively.} $\gamma$ is the minimum margin between the similarity of the correct caption and the unrelated caption. The loss function enables the sentence $X$ to be closer to the corresponding image $Y$ than the unrelated image $Z$. During the prediction time, the relationship between the text and images is calculated using the cosine similarity. 
	
	For general use, it is reasonable that some sentences, such as social constructs or metaphorical usage, are not paired with images after retrieval and have a low similarity score. In these cases, visual information might not be helpful. To measure how similar the retrieved images should be, we set a threshold $\delta$ to choose the top-ranked images for each sentence.

\section{Experiments}
\subsection{Task Settings}
Our evaluation is performed on the widely-used natural language generation and understanding tasks involving 14 NLP benchmark datasets that involve machine translation, natural language inference (NLI), semantic similarity, and text classification. Part of the NLU tasks is available from the GLUE benchmark \cite{wang2018glue}, which is a collection of nine NLU tasks.

\subsubsection{Neural Machine Translation}\label{sec:nmt}
Five widely-used translation tasks are used for model evaluation, including WMT'16 English-to-Romanian (En-Ro), WMT'14 English-to-German (En-De), WMT'14 English-to-French (En-Fr), and Multi30K dataset for WMT'16 and WMT'17, which are standard corpora for NMT and MMT evaluation.

(i) For the En-Ro task, we experiment with the officially provided parallel corpus: Europarl v7 and SETIMES2 from WMT'16 with 0.6M sentence pairs. We use \textit{newsdev2016} as the validation set and \textit{newstest2016} as the test set.

(ii) The En-De task has 4.43M bilingual sentence pairs of the WMT14 dataset used as training data, including Common Crawl, News Commentary, and Europarl v7. 
The \textit{newstest2013} and \textit{newstest2014} datasets are used as the validation set and test set, respectively.

(iii) The En-Fr task has 36M bilingual sentence pairs from the WMT14 dataset used as training data.
\textit{Newstest12} and \textit{newstest13} are combined for validation and \textit{newstest14} is used as the test set, following the setting of \cite{gehring-etal-2017-convolutional}. 

(iv) \textit{Multi30K} dataset contains 29K English$\rightarrow$\{German, French\} parallel sentence pairs with visual annotations. The 1,014 English$\rightarrow$\{German, French\} sentence pairs with visual annotations serve as the validation set. For WMT'16 and WMT'17 tasks, we have two test sets, test2016 and test2017, with 1,000 pairs for each.

\subsubsection{Natural Language Understanding}\label{sec:nlu}
The NLU task involves natural language inference, semantic similarity, and classification subtasks.

\noindent \textbf{Natural Language Inference}
involves reading a pair of sentences and assessing the relationship between their meanings, such as entailment, neutral, and contradiction. We evaluate the proposed method on four diverse datasets: SNLI \cite{Bowman2015A}, MNLI \cite{nangia2017repeval}, QNLI \cite{Rajpurkar2016SQuAD}, and RTE \cite{bentivogli2009fifth}.

\noindent \textbf{Semantic Similarity}
aims to predict whether two sentences are semantically equivalent. Three datasets are used: Microsoft Paraphrase Corpus (MRPC) \cite{dolan2005automatically}, Quora Question Pairs (QQP) dataset \cite{chen2018quora}, and Semantic Textual Similarity benchmark (STS-B) \cite{cer2017semeval}.

\noindent \textbf{Classification}
CoLA \cite{warstadt2018neural} is used to predict whether an English sentence is linguistically acceptable. SST-2 \cite{socher2013recursive} provides a dataset for sentiment classification that needs to determine whether the sentiment of a sentence extracted from movie reviews is positive or negative.




\subsection{Retrieval Setup}
This part describes the implementation of the image retrieval by the topic-image lookup table (TILT) and cross-modal retrieval model (CMRM):

\textbf{TILT}: 
We segment the sentences using the same BPE vocabulary as that for each source language. We select top-8 ($w=8$) high TF-IDF words, and the default number of images $m$ is set to 5.\footnote{In some cases when there is no paired image retrieved, we use the first-row all-zero vectors of the image lookup table as the ``blank features".} The detailed case study is shown in Section \ref{sec:sensitivity}. Image features are extracted from the averaged pooled features of a pre-trained ResNet50 CNN \cite{he2016deep}. The dimension of the feature maps is $V\in R^{2048}$.

\textbf{CMRM}: The cross-modal retrieval model is trained on the MS-COCO dataset \cite{lin2014microsoft}, which contains 123,287 images with five English captions per image. It is split into 82,783 training images, 5,000 validation images, and 5,000 test images. We use the Karpathy split \cite{karpathy2015deep} that forms 113,287 training, 5,000 validation and 5,000 test images. The model is implemented following the same settings as \citet{engilberge2018finding}, and produces state-of-the-art results (94.0\% R@10) for cross-modal retrieval. To ensure that each task can enjoy enough images, we set the similarity threshold $\delta$ to 0.4 and rank the paired images according to the similarity score. The maximum number of retrieved images $m$ for each sentence is set to eight according to our preliminary experiments. 

Multi30K and COCO datasets are used as the candidate seed image retrieval corpus for our downstream tasks.

\begin{table*}[t]
	\centering
		\caption{Results for the NMT tasks. ``++/+" after the BLEU score indicates that the proposed method (base: 5-8; large:9-12) was significantly better than the corresponding baseline Transformer (base or big) at significance level $p$ $<$0.01/0.05. 
	}
 \setlength{\tabcolsep}{12pt}
	{
		\begin{tabular}{lllrclrclr}
			\toprule
			\multirow{2}{*}{\#} &\multirow{2}{*}{\textbf{Model}}& \multicolumn{2}{c}{\textbf{En$\to$Ro}} &&  \multicolumn{2}{c}{\textbf{En$\to$De}} && \multicolumn{2}{c}{\textbf{En$\to$Fr}} \\ 
			\cmidrule{3-4} \cmidrule{6-7} \cmidrule{9-10}
			& & \textbf{BLEU} & \textbf{\#Param} && \textbf{BLEU} & \textbf{\#Param}  && \textbf{BLEU} & \textbf{\#Param} \\
			\midrule
			\multicolumn{10}{c}{\textit{Text-only Transformer}}                        \\ \midrule
			1 & Transformer-Base   & 32.66 & 61.54M&&  27.31  & 63.44M &&  38.52  & 63.83M   \\ 
		    2 &	Transformer-Big     &  33.85 & 207.02M&&  28.45 & 210.88M && 41.10  & 211.66M \\ 
			\midrule
			\multicolumn{10}{c}{\textit{Our MMT systems}}                             \\ \midrule
			3 & UVR-TILT$_\texttt{Multi30K}$ & 33.78++ &63.04M &&  \textbf{28.14}++ & 64.94M && 39.64++ & 65.33M     \\ 
			4 & UVR-TILT$_\texttt{COCO}$ & 34.08++ &63.04M &&   27.79+ &  64.94M &&  39.84++ & 65.33M     \\
			5 & UVR-CMRM$_\texttt{Multi30K}$ & 34.38++ &63.04M && 27.82+ & 64.94M & & 39.76++ & 65.33M     \\ 
			6 & UVR-CMRM$_\texttt{COCO}$ &\textbf{34.40}++ &63.04M &&  27.86+ & 64.94M && \textbf{40.24}++ & 65.33M     \\ 
			\midrule
			7 & UVR-TILT$_\texttt{Multi30K}$ & 34.46+ & 211.02M && 29.14++  & 214.89M   && \textbf{41.83}+ & 215.66M      \\ 
			8 & UVR-TILT$_\texttt{COCO}$ & 34.51+ & 211.02M &&  29.18++ & 214.89M   && 41.76+ & 215.66M      \\ 
			9 & UVR-CMRM$_\texttt{Multi30K}$ & 34.60++ &211.02M && 28.96+  & 64.94M && 41.79+ & 215.66M    \\
			10 & UVR-CMRM$_\texttt{COCO}$ &\textbf{34.62}++ &211.02M &&  \textbf{29.21}++ & 64.94M && 41.82+ & 215.66M    \\ 
			\bottomrule
		\end{tabular}
	}
		\label{tbl:nmt}
		

\end{table*}

\begin{table*}[htb]
 \setlength{\tabcolsep}{12.6pt}
\centering
	\caption{Results (BLEU) from the test2016 and test2017 for the MMT task. ``++/+" after the BLEU score indicates that the proposed method (base: 5-8; large: 9-12) was significantly better than the corresponding baseline Transformer (base or big) at significance level $p$ $<$0.01/0.05. } 
	{
		\begin{tabular}{llllrlllr}
			\toprule
            \multirow{2}{*}{\#}    & \multicolumn{1}{c}{\multirow{2}{*}{\textbf{Model}}} & \multicolumn{3}{c}{\textbf{En-De}} &   & \multicolumn{3}{c}{\textbf{En-Fr}}    \\ 
            \cmidrule{3-5} \cmidrule{7-9}
			& \multicolumn{1}{c}{} & \textbf{Test2016} & \textbf{Test2017} & \textbf{\#Param} & &\textbf{Test2016} & \textbf{Test2017} & \textbf{\#Param} \\ 
			\midrule
			\multicolumn{8}{c}{\textit{Text-only Transformer}} \\ \midrule
		    1 &   Transformer-Base     &    35.59      &     26.31     &   49.15M    &  &    57.88      &   48.55     &    49.07M     \\
			2 &   Transformer-Big    &   36.86       &    27.62      &   186.38M    &  &  56.97        &    48.17      &    186.23M     \\
			\midrule
			\multicolumn{8}{c}{\textit{Standard MMT systems}} \\ \midrule
		    3 &  MMT-Base   &      35.09     &     27.10    &   50.72M    &  &    57.40       &    48.02     &       50.65M  \\  
			4 &   MMT-Big &    35.60       &    28.02     &     190.58M      &    & 57.87       &   49.63   &     190.43M    \\ 
			\midrule
			\multicolumn{8}{c}{\textit{Our MMT systems}} \\ 
			\midrule
		    5 & UVR-TILT$_\texttt{Multi30K}$  & 35.72 &    26.87+    &    50.72M    &  &    58.32+      &   48.69    &   50.65M \\
			6 & UVR-TILT$_\texttt{COCO}$  & 35.67 &   26.89+   &    50.72M     & &    58.21+     &  48.73 &   50.65M \\
		    7 & UVR-CMRM$_\texttt{Multi30K}$  &   \textbf{36.38}+ &   \textbf{27.34}++    &    50.72M     & &    \textbf{58.53}+   &  \textbf{49.28}+  &   50.65M \\
		    8 & UVR-CMRM$_\texttt{COCO}$  &  35.78   &  26.92+     &    50.72M     & &    58.46+    &  48.58    &   50.65M \\
			\midrule
			9 & UVR-TILT$_\texttt{Multi30K}$ &   37.02        &   28.63++       &    190.58M     & &     57.53+    &     48.46     &    190.43M     \\
			10 & UVR-TILT$_\texttt{COCO}$ &    36.94  &   28.69++        &    190.58M     &&   57.62+     &   48.39  &    190.43M     \\
			11 &  UVR-CMRM$_\texttt{Multi30K}$ & 37.16  &  \textbf{28.82++} &    190.58M   &  &  \textbf{58.37}++ & \textbf{48.77}+  &    190.43M     \\
	    	12	&  UVR-CMRM$_\texttt{COCO}$ & \textbf{37.28}+ & 28.71++ &    190.58M   &  & 57.60+ & 48.42 &    190.43M     \\
			\bottomrule
		\end{tabular}
	}
\label{tbl:mmt}

\end{table*}

\begin{table*}[htb]
 \setlength{\tabcolsep}{14.6pt}
\centering
	\caption{Comparison with public methods on the Multi30K MMT dataset. The results of existing methods are from \cite{wu-etal-2021-good}. ``++/+" after the BLEU score indicates that the proposed method was significantly better than the corresponding baseline Transformer (tiny) at significance level $p$ $<$0.01/0.05. } 
	{
		\begin{tabular}{llllrlllr}
			\toprule
            \multirow{2}{*}{\#}    & \multicolumn{1}{c}{\multirow{2}{*}{\textbf{Model}}} & \multicolumn{3}{c}{\textbf{En-De}} &   & \multicolumn{3}{c}{\textbf{En-Fr}}    \\ 
            \cmidrule{3-5} \cmidrule{7-9}
			& \multicolumn{1}{c}{} & \textbf{Test2016} & \textbf{Test2017} & \textbf{\#Param} & &\textbf{Test2016} & \textbf{Test2017} & \textbf{\#Param} \\ 
			\midrule
			\multicolumn{9}{c}{\textit{Text-only Transformer}} \\ \midrule
		    1 &   Transformer-Tiny    &   40.38    &  32.86   &  2.6M  &  &    61.00 & 52.42   &    2.6M    \\
			\midrule
			\multicolumn{8}{c}{\textit{Existing MMT systems}} \\ 
			\midrule
			2 & GMNMT \cite{yin2020novel} &  39.8 & 32.2   & 4.0M & &    60.9 & 53.9   &  - \\
			3 & DCCN \cite{lin2020dynamic} & 39.7 & 31.0   &  17.1M & &    61.2 & 54.3 &   16.9M \\
			\midrule
			\multicolumn{9}{c}{\textit{Our MMT systems}} \\ 
			\midrule
			4 & UVR-TILT$_\texttt{Tiny}$ &    \textbf{41.27}++  &  \textbf{33.62}++      &  2.9M       &   &     \textbf{61.60}+   &      \textbf{54.83}++      &     2.9M       \\
			5 &  UVR-CMRM$_\texttt{Tiny}$ &    40.94+     &   33.11+   &  2.9M    &  & 61.50+ &  53.64++ &   2.9M         \\
			\bottomrule
		\end{tabular}
	}
\label{tbl:mmt_public}

\end{table*}

\begin{table*}[htb]
	\centering
	\setlength{\tabcolsep}{7.0pt}
		\caption{{Test results on the GLUE benchmark. The best results are marked in boldface.}}
	\label{GLUE}
	{
	\begin{tabular}{llccccccccccccc}  
	\toprule
	\multirow{2}{*}{\#}    & \multicolumn{1}{c}{\multirow{2}{*}{\textbf{Model}}} & \multicolumn{2}{c}{\textbf{Classification}} && \multicolumn{3}{c}{\textbf{Semantic Similarity}} && \multicolumn{4}{c}{\textbf{Language Inference}} && \multirow{2}{*}{\textbf{Average}}  \\
	\cmidrule{3-4}\cmidrule{6-8}\cmidrule{10-13}
	&	&\textbf{CoLA} & \textbf{SST-2} && \textbf{MRPC} & \textbf{STS-B} & \textbf{QQP} && \textbf{MNLI} &\textbf{QNLI} & \textbf{RTE} & \textbf{SNLI} && \\
\midrule
\multicolumn{11}{c}{\emph{Public Systems}} \\
\midrule
1 & BERT \cite{devlin2018bert} &60.5 &94.9 && 85.4 &87.6 &89.3 & &86.7  &92.7 &70.1 & - && 83.4\\
2 & MT-DNN \cite{liu2019multi} &62.5 &95.6 && 88.2 &89.5 &89.6 & &86.7  &93.1 &81.4 & 91.6 && 86.4\\
3 & BERT + Voken-cls \cite{tan2020vokenization} & - & 92.2 && - & - & 88.6 && 82.6 & 88.6 & - & - && - \\ 
4 & UniT \citep{hu2021unit} & - & 91.5 && - & - & 88.4 && 79.8 & 88.0 & - & - & - & - \\
\midrule
\multicolumn{11}{c}{\emph{Our Systems}} \\
\midrule
5 & Baseline (BERT$_\text{WWM}$)   & \textbf{63.6} & 93.6 && 87.0 & 90.2 &88.8 && 87.2 &93.9 & 77.3 &91.6 && 85.9\\
6 & UVR-TILT$_\texttt{Multi30K}$ & 62.5 & 94.7  && 87.7 & 89.8 & 89.4 && 87.2 & 94.1 & \textbf{84.5} & \textbf{91.7} && 86.8  \\
7 & UVR-TILT$_\texttt{COCO}$ & 62.8 & \textbf{94.9} && 87.4 & 90.2 & 89.7 && 86.9 & 94.0 & 83.6 & \textbf{91.7} && 86.8\\
8 & UVR-CMRM$_\texttt{Multi30K}$ & 63.0 & 94.3 && 87.8& 90.2 & 89.6  && 87.3 & 93.8& 83.8& 91.6 && 86.8 \\
9 & UVR-CMRM$_\texttt{COCO}$ & 63.2 & 94.6 && \textbf{87.9} & \textbf{90.3} & \textbf{89.8} && \textbf{87.4} & \textbf{94.2} & 83.9 & \textbf{91.7} && 87.0 \\
	\bottomrule
	\end{tabular}
	}

\end{table*}

\subsection{Model Implementation}\label{baseline}
Since our task involves text generation and understanding, we have two kinds of baselines, the NLG model for translation and the NLU model for the other tasks.
\subsubsection{NLG Model}
Our baseline for NLG is encoder-decoder Transformer \cite{NIPS2017_7181}. We use six layers for the encoder and the decoder. The number of dimensions of all input and output layers is set to 512 and 1024 for \textit{base} and \textit{big} models. For MMT experiments on the Multi30K dataset, we also use the tiny setting where the dimension of the input and output layer is 128. The inner feed-forward neural network layer is set to 2048. The heads of all multi-head modules are set to eight in both the encoder and decoder layers. For the \textit{Multi30K} dataset, we further evaluate a multimodal baseline (denoted as MMT) where each source sentence was paired with an original image. The other settings were the same as our proposed model.

The byte pair encoding algorithm is adopted to segment sentences into subword sequences, with the vocabulary size set to 40,000.
In each training batch, a set of sentence pairs contains approximately 4096$\times$4 source tokens and 4096$\times$4 target tokens. 
During training, the value of label smoothing is set to 0.1, and the attention dropout and residual dropout rates are \textit{p} = 0.1. 
We used Adam optimizer~\cite{DBLP:journals/corr/KingmaB14} to tune the parameters of the model.
The learning rate is varied under a warm-up strategy with 8,000 steps.
For evaluation, we validate the model with an interval of 1,000 batches on the validation set.
For the \textit{Multi30K} dataset, we train the model up to 10,000 steps, and the training will be early-stopped if the validation set BLEU score does not improve for ten epochs. 
For the En-De, En-Ro, and En-Fr tasks, following the training of 200,000 batches, the model with the highest BLEU score of the validation set is selected to evaluate the test sets.
During the decoding, the beam size is set to five. Multi-bleu.perl is used to compute case-sensitive $4$-gram BLEU scores for all test sets.\footnote{\url{https://github.com/moses-smt/mosesdecoder/tree/RELEASE-4.0/scripts/generic/multi-bleu.perl}.}
We follow the model configurations of \cite{NIPS2017_7181} to train big models for WMT En-Ro, En-De, and En-Fr translation tasks. The experiments of NLG are conducted with \textit{fairseq} \cite{ott2019fairseq}.\footnote{\url{https://github.com/pytorch/fairseq}.} 

{For the statistical tests, we perform the paired bootstrap resampling test  \cite{koehn2004statistical} to measure the reliability of the conclusion that our system is better than the baseline. Our implementation is based on the public toolkit.\footnote{\url{https://github.com/neubig/util-scripts/blob/master/paired-bootstrap.py}} Two thousand bootstrap samples are used for each significance test.}

\subsubsection{NLU Model}
For the NLU tasks, the baseline is encoder-only BERT \cite{devlin2018bert}.\footnote{\url{https://github.com/huggingface/transformers}.} We use the whole-word-mask (WWM) version of the pre-trained weights due to its more stable, reproducible, and slightly better performance than the original large version \cite{devlin2018bert}. The initial learning rate is set in the range \{2e-5, 3e-5\} with a warm-up rate of 0.1 and L2 weight decay of 0.01. The batch size is selected from \{16, 24, 32\}. The maximum number of epochs is set in the range [2, 5]. Texts are tokenized using SentencePiece,\footnote{\url{https://github.com/google/sentencepiece}.} with a maximum length of 128.

\subsection{Main Results}\label{sec:result}

Tables~\ref{tbl:nmt}-\ref{GLUE} show the results for the 14 NMT, MMT, and NLU tasks, respectively. According to the results, we have the following observations:

(i) According to the machine translation results in Tables 1-2, the proposed UVR methods significantly outperform the baselines according to the statistical test, demonstrating the effectiveness of modeling visual information for text-only NMT. In particular, the superiority is observed in the translation tasks of three language pairs with different training data scales, verifying that the proposed approach is a universal method for improving translation performance.

(ii) Our method introduces only 1.5M and 4.0M parameters for the base and big Transformers, respectively. The number is less than 3\% of the baseline parameters as we use the fixed image embeddings from the pre-trained ResNet feature extractor. Besides, the training time is basically the same as the baseline model (Section \ref{time}).

(iii) Results in Tables \ref{tbl:mmt}-\ref{tbl:mmt_public} show that our model can generally outperform the Transformer baseline in multimodal settings that could benefit from the gold sentence-image annotations. Compared with the results in text-only NMT, we find that the image enhancement sometimes gives marginal contribution, which is consistent with the findings in previous work \cite{zhang2017nict,gronroos2018memad,caglayan2019probing}. 
The most plausible reason might be that the sentences in \textit{Multi30K} are quite simple, short, and repetitive, so that the source text itself is sufficient to perform the translation \cite{caglayan2019probing,ive2019distilling}. We also see that the big models sometimes show inferior results. The possible reason is that the dataset is too small to effectively train such big models, which easily suffer from over-fitting issues. The hypothesis is also supported by our superior results with the tiny model setting in Table \ref{tbl:mmt_public}. The observation verifies our assumption of the current bottleneck of MMT due to the limitation of \textit{Multi30K} and shows the necessity of our new methodology of transferring multimodality into more standard and mature text-only NMT tasks.

\begin{figure}
\setlength{\abovecaptionskip}{0pt}
\begin{center}
\pgfplotsset{height=5.2cm,width=7cm,compat=1.14,every axis/.append style={thick},every axis legend/.append style={ at={(0.95,0.95)}},legend columns=1 row=2} 

\begin{tikzpicture} \tikzset{every node}=[font=\small] 
\begin{axis}[hide x axis,
                 axis y line*=right,
                 axis x line*=right,
                 ybar=5pt,
                 width=7cm,
                 legend pos=north west,
                 bar width=6pt,
                 enlargelimits=0.13,
                 ylabel={Percentage},
                 font=\small,
                 ymax= 0.55,
                 symbolic x coords={0,1,2,3,4,5,6,7,8},
                 xtick=data,
                 bar width=9pt,
                 ],
                 \addlegendimage{/pgfplots/refstyle=plot_line}\addlegendentry{D-value}
                \addplot+[teal] coordinates {(0,0.351283825
) (1,0.404035945
) (2,0.445153362
) (3,0.366086885
) (4,0.456889273
) (5,0.370326339
) (6,0.421974013
) (7,0.449577652
) (8,0.44066048
)};
                \addlegendentry{\small Coverage}
                 \end{axis}
                 
                 \begin{axis} [width=7cm,enlargelimits=0.13, xticklabels={CoLA, SST-2, MRPC, STS-B, QQP, MNLI, QNLI, RTE, SNLI}, axis y line*=left, axis x line*=left, xtick={0,1,2,3,4,5,6,7,8}, x tick label style={rotate=45},
  ylabel={Accuracy},
  ymax=8.8,
  ylabel style={align=left},font=\small]
\addplot+ [smooth, mark=square*,mark size=1.2pt,mark options={mark color=red}, color=red] coordinates {(0,-1.1) (1,1) (2,0.7) (3,-0.4) (4,0.6) (5,0) (6,0.2) (7,7.2) (8,0.1)};
\label{plot_line}
\end{axis}

\end{tikzpicture}
\end{center}
  \caption{Accuracy difference between our method and baseline compared with the coverage percentage of tokens that can be paired with images in each dataset.}
  \label{overlap_task}
\end{figure}
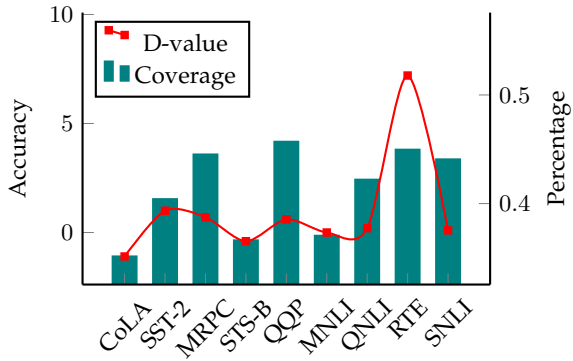

\begin{table}[t]
\setlength{\tabcolsep}{8.8pt}
\centering
	{
		\caption{{Validation results on GLUE datasets in different sizes: small datasets with less than 10$k$ examples (RTE and STS-B), and a large dataset with more than 10$k$ examples (QNLI). The MT-DNN results are reproduced using the released weights \cite{liu2019multi}. }}
		\label{tbl:mt-dnn}} 
	{
	\begin{tabular}{lccc}
		\toprule
		\textbf{Model} &  \textbf{RTE} & \textbf{STS-B} & \textbf{QNLI} \\
		\midrule
        MT-DNN$_\textup{base}$ & 78.94\scriptsize{$\pm$0.83} & 88.14\scriptsize{$\pm$0.40} & 90.32\scriptsize{$\pm$0.12}\\
        \quad w/ UVR-TILT & 81.59\scriptsize{$\pm$0.63} & 90.16\scriptsize{$\pm$0.07} & 91.31\scriptsize{$\pm$0.19}\\
        \midrule
        MT-DNN$_\textup{large}$ & 78.70\scriptsize{$\pm$1.65} & 90.16\scriptsize{$\pm$0.17} & 92.04\scriptsize{$\pm$0.27}\\
        \quad w/ UVR-TILT & 82.19\scriptsize{$\pm$1.37} & 91.32\scriptsize{$\pm$0.12} & 92.78\scriptsize{$\pm$0.12}\\
		\bottomrule
	\end{tabular}
	}
\end{table}

(iv) Table~\ref{GLUE} shows our method is generally helpful for a wide range of NLU tasks in the GLUE benchmark, which verifies the effectiveness of modeling visual information for language understanding. {We are interested in whether public methods, such as MT-DNN, can be further enhanced by our method, we apply our UVR-TILT method to the MT-DNN model based on the same implementation in BERT. According to the results in Table \ref{tbl:mt-dnn}, both the base and large models are enhanced, and we observe consistent gains in different datasets, especially the small datasets.} For the results in Tables \ref{GLUE}-\ref{tbl:mt-dnn}, we notice a few inferior or marginally better performances in the CoLA, MNLI, and QNLI tasks. We calculate the accuracy difference (D-value) between our method and baseline compared with the coverage percentage of tokens that can be paired with images in each dataset using UVR-TILT$_\texttt{Multi30K}$. From Figure \ref{overlap_task}, we see that those datasets are commonly paired with a relatively small number of images so that the visual signals can enhance only a small proportion of token representations. In addition, some datasets, i.e., ColA, mainly require linguistic knowledge for solving the tasks, so introducing visual modality might not benefit such tasks, which corresponds to the common shortcoming of vision injection for language tasks. For MNLI and QNLI, the possible reason for the marginal improvements would be that both of the datasets are quite large. Still, the task is relatively simple, so the model might well solve the tasks directly via the text representations. In this scenario, the visual features might only provide the regularization effect to improve the model robustness \cite{brown2003use,noh2017regularizing,brownlee2019train}.

(v) For the two retrieval methods, we observe that \textit{UVR-CMRM} is slightly better than \textit{UVR-TILT} in general, and the results of the two methods are pretty close for the MMT task. We find a performance tradeoff between them: there might be more accurate similarity calculation after cross-modal pre-training than the direct topic extraction. However, controlling the proper similarity threshold would be a heuristic for each dataset. In contrast, the advantage of \textit{UVR-TILT} is the simple preprocessing, which only requires TF-IDF-based topic extraction and matching.

(vi) We also compare the results using different seed corpora, i.e., Multi30K and COCO. For the MMT task in Table \ref{tbl:mmt}, using Multi30k is basically similar to or slightly better than COCO in general, as the task could enjoy the images in the same domain. For the out-of-domain evaluations in Table \ref{tbl:nmt} and Table \ref{GLUE}, we obverse that using COCO generally achieves better results because the size of COCO images is three times that of Multi30K, which could provide more diverse image features.

\begin{figure}[t]
	\centering
	\includegraphics[width=0.42\textwidth]{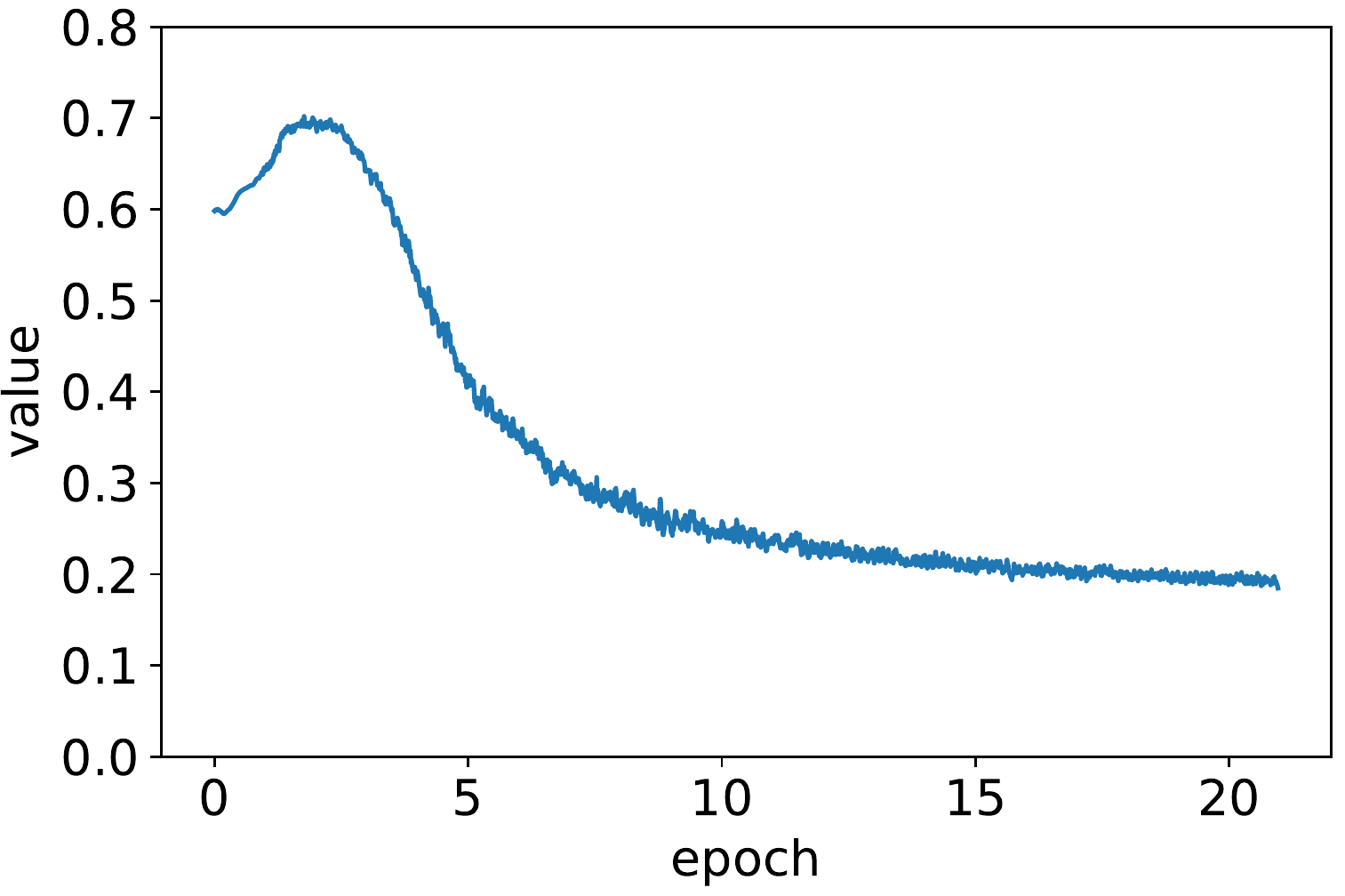}
	\caption{\label{fig:gate_values} {Illustration of the gate values $\lambda$ with the UVR-TILT method on Multi30K En-De Test2016.}}
\end{figure}

\section{Analysis}\label{ana}
{This section presents our exploration on the role of visual contexts, which involves two aspects, when the visual context helps and how the visual context helps language representations.} In the following analysis part, we use Multi30K for UVR-TILT and COCO for UVR-CMRM by default.

\subsection{Dynamics of the visual information}\label{sec:role}
{To explore the role of visual context in the training process, we illustrate the gate values $\lambda$ (defined in Eq. \ref{eq3:Lamda_Image}) in Figure \ref{fig:gate_values} where a larger value indicates more dependence on the visual context in the fusion process. We observe that the model relies on the visual context in the early stages, and the role of visual information changes dynamically across training. When the training starts, the model accommodates the visual information to a large extent ($\lambda\geq0.6$), indicating that the model tends to trust the visual context, which could provide useful information in the early stages. With more knowledge captured as the training continues, the contributions of the visual contexts appeal to decay. In the following parts, we will further discuss the specific effectiveness of visual representations in language modeling.}

\begin{figure*}[htb]
	\centering
	\includegraphics[width=0.98\textwidth]{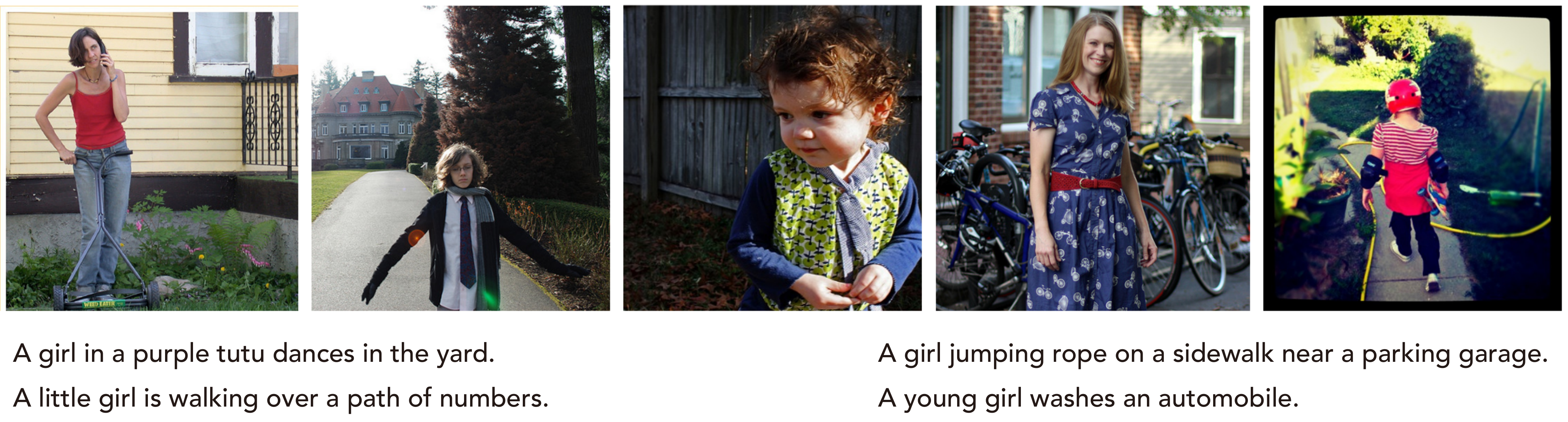}
	\caption{Examples of sentences that share the same retrieved images, in which the common topic is about ``girl". Sentences with similar topics tend to be paired with similar or even the same images, and vice versa. This means that images may provide topic information, which benefits the modeling of similar sentences.}
	\label{same}
\end{figure*}
    
\subsection{Pairwise relationship across modalities}\label{sec:benefits}
The benefits of the universal representation method could be two folds: (i) the content connection of the sentences and images; (ii) the topic-aware co-occurrence of similar images and sentences. 
According to Distributional Hypothesis \cite{harris1954distributional}, which states that \textit{words that occur in similar contexts tend to have similar meanings}, we are inspired to extend the concept in the multimodal world, \textit{the sentences with similar meanings would be likely to pair with similar even the same images}. Therefore, the consistent images (with a related topic) could play the role of topic or type hints for similar sentence modeling.

After using our image retrieval method, sentences with similar topics tend to be paired with similar or even the same images, and vice versa. Figure \ref{same} shows examples in which the common topic is about ``girl". This means that images may provide topic information, which benefits the modeling of similar sentences. Thus, aside from the image embeddings' inner meaning (vectors), there is a mapping relationship between the sentence and images after the retrieval process.

For the image embeddings, as described in Section \ref{sec:image_encoder}, we adopt the embedding lookup to fetch the embedding features for each image, which is very similar to the way of using word embedding by treating each image as a ``word". The weights of the embedding features are derived from the average pooled output of ResNet, where each image is represented as a 2400-d vector.
For all the 29,000 images (e.g., using Multi30K), we have an embedding layer with size (29000, 2400). The ``content" of the image can be seen as embedding initialization. The pre-initialized embedding weights might yield slight improvement gains. However, the neural network can also be effectively trained with random initialization \cite{neishi2017bag,kocmi2017exploration}. In contrast, whether to use the embedded feature is more critical. 
In other words, the mapping relationship of the sentences and images in image embedding would be essential, i.e., similar sentences (with the same topic words) tend to map the same or similar image after the word-image lookup process.

To verify the hypotheses, we conduct the following ablations: we replace the ResNet50 feature extractor in our UVR-TILT model with (1) \emph{ResNet101} and (2) \emph{ResNet152}; additionally, we compare the results with the following operations: (3) \emph{Shuffle}: shuffle the image features but retain the lookup table; (4) \emph{Random Init}: randomly initialize the image embedding but keep the lookup table; (5) \emph{Random Mapping}: randomly retrieve unrelated images.

\begin{table}[t]
	\centering
	\caption{{Ablation for the image embedding operation on En-Ro. The scores are reported by means and standard deviations for three random seeds.}}
\label{tbl:embedding}
	\setlength{\tabcolsep}{13.8pt}
	{%
		\begin{tabular}{lc}
			\toprule
			\textbf{Method} & \textbf{BLEU Score} \\
			\midrule
			Baseline & 32.75\scriptsize{$\pm$0.10} \\
			UVR-TILT & 33.72\scriptsize{$\pm$0.08} \\
			\quad w/ (1) Res101 & 33.65\scriptsize{$\pm$0.06} \\
			\quad w/ (2) Res152 & 33.82\scriptsize{$\pm$0.07} \\
			\quad w/ (3) Shuffle & 33.40\scriptsize{$\pm$0.14} \\
			\quad w/ (4) Random Init & 33.08\scriptsize{$\pm$0.17} \\
			\quad w/ (5) Random Mapping & 32.05\scriptsize{$\pm$0.14} \\
			\bottomrule
	\end{tabular}}

\end{table}

\begin{figure*}[htb]
   \centering
	\includegraphics[width=0.9\textwidth]{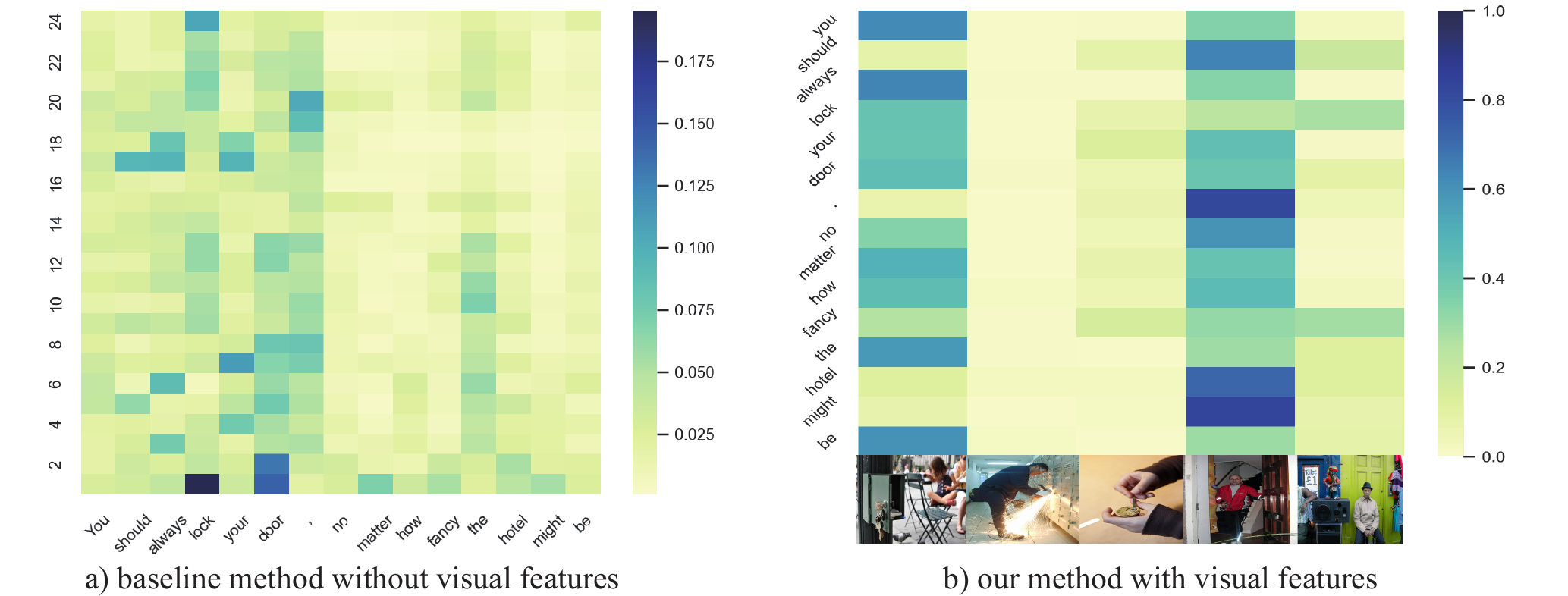}
	\caption{{{Visualization of (a) attention weights of the input tokens with regards to the probed token ``lock'' across different layers (Y-axis) using the BERT baseline; (b) image-to-word attention from our model. The illustration shows that the images provide fine-grained grounding information about the relationship between concepts and events, e.g., ``lock", ``door", ``fancy", ``hotel".}}
	}
	\label{fig:vis-1}
\end{figure*}

{Table \ref{tbl:embedding} shows the ablation results. The BLEU scores of models 1-4 are close to the proposed UVR method, and those ablated methods still outperform the baseline, indicating that using image features generally yields better performance than the baseline. In addition, either replacing the trained image features (model 4) or disturbing the mapping information (model 5) leads to a performance drop ($\downarrow$0.64/$\downarrow$1.67, respectively), which indicates that both the image features and mapping information are contributing factors. Compared with image features, the mapping information has a larger impact, which verifies our prior hypothesis that the consistent images with a related topic could play the role of topic or type hints for similar sentence modeling in the whole training process. With the mapping information, the same images will be assigned to the same context. During training, the image features will be learned just like word embeddings. Therefore, it does not mean that the image features are not very helpful, but the mapping information in the lookup table reduces the dependence on the trained image features.} 
 

From the view of an individual image, image content (embedding) has an effect. If we maintain the pairwise relationship between the sentence and image, the result is still higher than the baseline, even with shuffled or random image embeddings. This indicates that the pairwise relationship is a vital contributor. From the macro perspective of sentence-image co-occurrence, image information plays the role of topic information, where similar sentences tend to pair with similar images. The observation corresponds to the distributional hypothesis. This explains the potential effects of the pairwise relationship.

This finding may potentially facilitate future research because most existing studies focus on the content of the individual image itself. We highlight the pairwise relationship across modalities as a different research line to bridge the gap between language and image modeling.


\subsection{Handling incomplete source texts}
Content words are naturally related to specific content, such as car, room, play. We collect a list of tokens that have more than ten occurrences in the Multi30K training set after removing all stop words following \cite{tan2020vokenization}. We remove those tokens in the source sentence, which occupy 42.87\% of tokens in the token dictionary. Table \ref{tbl:incomplete} shows the results of the baseline model and our model with the incomplete source texts. We observe that our model achieves noticeable gains over the baseline. {The results verify that the visual representation can reduce the gap of the missing information from the content words in the source texts.}

\begin{table}[t]
\centering
	\caption{{Results of MMT with incomplete source texts by removing visually grounded tokens in the Multi30K dataset. The scores are reported by means and standard deviations for three random seeds.}} 
	{
		\begin{tabular}{llllll}
			\toprule
             \multicolumn{1}{c}{\multirow{2}{*}{\textbf{Model}}} & \multicolumn{2}{c}{\textbf{En-De}} &   & \multicolumn{2}{c}{\textbf{En-Fr}}    \\ 
            \cmidrule{2-3} \cmidrule{5-6}
			\multicolumn{1}{c}{} & \textbf{Test2016} & \textbf{Test2017}  & &\textbf{Test2016} & \textbf{Test2017} \\ 
			\midrule
		     Baseline   & 10.94\scriptsize{$\pm$0.21}  &  7.75\scriptsize{$\pm$0.24}   &   &  18.61\scriptsize{$\pm$0.16}  &  15.01\scriptsize{$\pm$0.15}       \\
			 UVR-TILT  &  12.80\scriptsize{$\pm$0.18}   & 9.15\scriptsize{$\pm$0.19}  &   & 19.60\scriptsize{$\pm$0.17}  &  15.77\scriptsize{$\pm$0.20}   \\
			\bottomrule
		\end{tabular}
	}
\label{tbl:incomplete}
\end{table}

\subsection{Knowledge grounding with the visual context}
{To gain an insight into the process of multimodal integration by our model, we analyze the attention distributions ($\alpha$ in Eq.\ref{eq2:Image_Representation}) at the multimodal integration layer. {Figure \ref{fig:vis-1} shows the attention distributions of (a) the baseline and (b) our model \footnote{Our model is the UVR-TILT trained on the CoLA dataset.} for an example randomly selected from our GLUE validation sets, ``\textit{You should always lock your door, no matter how fancy the hotel might be}.'' For comparison, the baseline is implemented following \citet{abnar2020quantifying}. Concretely, we collect the attention weights of all the input tokens with regards to a targeted token ``lock'' across different layers (i.e., \{2, 4, \dots, 24\}). As the baseline uses the representation of the last layer for prediction, we focus on the attention distributions of the last layer.}

{Compared with the baseline that only captures partial relations in the last layer, e.g., with a lack of relationship among \{``lock", ``door", ``hotel''\}, our model provides more fine-grained connections. In detail, two generic patterns are observed in these examples.}} 

{(i) the images appear to match the concepts and actions with the texts, in other words, the images tend to provide fine-grained grounding information about the relationship between concepts and events, e.g., \{"lock", "door", "fancy", "hotel"\}. }

{(ii) our model can resist irrelevant information from noisy images. For example, the second and third images yield low attention scores for the texts.}

\begin{table}[t]
\setlength{\tabcolsep}{12pt}
\centering
	{
		\caption{{Results (BLEU score) of the multimodal disambiguation experiments on WAT'19 English to Hindi dataset. The scores are reported by means and standard deviations for three random seeds.}}
		\label{tbl:disam}} 
	{
	\begin{tabular}{lccc}
		\toprule
		\textbf{Model} &  \textbf{Validation} & \textbf{Test} & \textbf{Challenge} \\
		\midrule
        Baseline & 47.04\scriptsize{$\pm$0.27} & 39.33\scriptsize{$\pm$0.24} & 20.52\scriptsize{$\pm$0.14}\\
        UVR-CMRM & 47.49\scriptsize{$\pm$0.06} & 39.81\scriptsize{$\pm$0.12} & 21.62\scriptsize{$\pm$0.08}\\
		\bottomrule
	\end{tabular}
	}

\end{table}

\begin{figure*}[htb]
   \centering
	\includegraphics[width=0.9\textwidth]{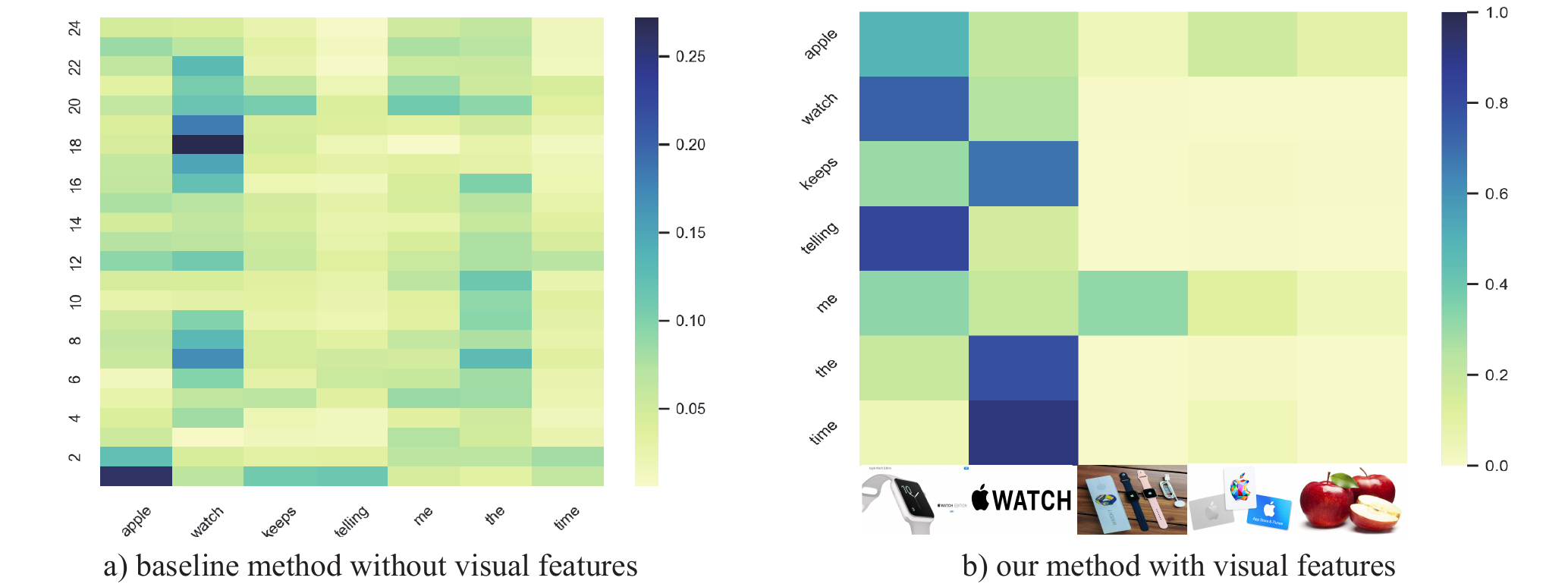}
	\caption{{Visualization of (a) attention weights of the input tokens with regards to the ambiguous token ``apple'' across different layers (Y-axis) using the BERT baseline; b) image-to-word attention from our model. The illustration shows that the images bridge the connection between ``apple'', ``watch'', ``time'', helping disambiguate the meaning of ``apple''.}
	}
	\label{fig:vis-2}
\end{figure*}

\begin{table*}[htb]

\centering
	{
		\caption{Selected eight probing tasks \cite{conneau2018you} to study what syntactic and semantic properties are captured by the encoders.}
		\label{tbl:prob_type}} 
	{
	{
\begin{tabular}{p{1.5cm}|p{1.5cm}|p{13cm}}
		\hline
		
		\hline
		\multicolumn{2}{c|}{\textbf{Probing Tasks}} &  \makecell[c]{\textbf{Content}} \\ 
		\hline
        \multirow{3}{*}{Syntactic} & TrDep & Checking whether an encoder infers the hierarchical structure of sentence \\
        \cline{2-3}
        & \multirow{2}{*}{ToCo} & Sentences should be classified in terms of the sequence of top constituents immediately below the sentence node \\
        \cline{2-3}
        & BShif & Testing whether two consecutive tokens within the sentence have been inverted \\
        \hline
        \multirow{5}{*}{Semantic} & Tense & Asking for the tense of the main clause verb \\
        \cline{2-3}
        & SubN & Focusing on the number of the main clause’s subject\\
        \cline{2-3}
        & ObjN & Testing for the number of the direct object of the main clause \\
        \cline{2-3}
        & \multirow{2}{*}{SoMo} & Some sentences are modified by replacing a random noun or verb with another one and the classifier should tell whether a sentence has been modified  \\
        \cline{2-3}
        & CoIn & Containing sentences made of two coordinate clauses \\
		\hline
		
		\hline
	\end{tabular}
	}
	}

\end{table*}

\begin{table*}[htb]
\setlength{\tabcolsep}{14.2pt}
\centering
	{
		\caption{Classification accuracy on eight probing tasks of evaluating linguistics embedded in the encoder outputs.}
		\label{tbl:prob_result}} 
	{
	\begin{tabular}{lllllllllll}
		\toprule
		\multirow{2}{*}{\textbf{Model}} & \multicolumn{3}{c}{\textbf{Syntactic}} && \multicolumn{5}{c}{\textbf{Semantic}} \\
		\cmidrule{2-4}\cmidrule{6-10}
		& \textbf{TrDep} & \textbf{ToCo} & \textbf{BShif} && \textbf{Tense} & \textbf{SubN} & \textbf{ObjN} & \textbf{SoMo} & \textbf{CoIn} \\
		\midrule
        Baseline & 28.34  & 58.33 & 76.34 && 80.66 & 72.02 & 68.57 & 64.42 & 67.51 \\
        UVR-CMRM & 28.53 & 58.64 & 77.72 && 80.97 & 73.79 & 69.66 & 65.44 & 67.23 \\
		\bottomrule
	\end{tabular}
	}
\end{table*}

\subsection{Disambiguation}\label{sec:Disambiguation}
A natural intuition of using visual clues for text representation is the advantage of alleviating the ambiguation of language. To evaluate the model performance for disambiguation, we use a dataset from the HVG \cite{parida2019hindi}, which serves as a part of the WAT'19 Multimodal Translation Task.\footnote{http://lotus.kuee.kyoto-u.ac.jp/WAT/WAT2019/index.html} The dataset consists of a total of 31525 randomly selected images from Visual Genome \cite{krishna2017visual} and a parallel image caption corpus in English-Hindi for selected image segments. The training part consists of 29$K$ English and Hindi short captions of rectangular areas in photos of various scenes, and it is complemented by three evaluation subsets: validation, test, and challenge test set (Challenge).
The challenge test set is created by searching for (particularly) ambiguous English words based on the embedding similarity and manually selecting those where the image helps to resolve the ambiguity. We do not use the images but follow the same settings as the experiments on Multi30K. As the results shown in Table \ref{tbl:disam}, we observe that our UVR model works effectively on the challenge disambiguation set, indicating that the visual information induced by retrieved images allows disambiguation of translation.

{Figure \ref{fig:vis-2} shows a heatmap of the attention visualization on an ambiguous sentence, ``apple watch keeps telling me the time". In Figure \ref{fig:vis-2}(a), the baseline model fails to capture the relationship between ``apple'' with ``time''. In Figure \ref{fig:vis-2}(b), the retrieved images bridge the connection among ``apple'', ``watch'' and ``time'', which helps disambiguate the meaning of ``apple".}

\subsection{Linguistic Analysis}\label{sec:linguisctic}
We are interested in what knowledge is learned in the universal representations. In this section, we select eight widely-used language probing tasks \cite{conneau2018you} (see Table \ref{tbl:prob_type}) to study what kind of syntactic and semantic properties are captured by the encoders. Specifically, we use the encoders of the baseline BERT-based SNLI model,\footnote{We select the SNLI model because NLI models show good generalization capacity for language representation \cite{zhang2019semantics,clark2019electra}, which is supposed to be a strong test-bed for the evaluation.} and our UVR-CMRM visual representation model to generate the sentence representations of input, which are used to carry out the above eight probing tasks. The results are as shown in Table \ref{tbl:prob_result}.

Concerning semantic properties, our model gains
the most significant improvement on the \textit{SubN} and \textit{ObjN} tasks. The result indicates that visual information helps NMT to identify and represent the subject and object information, which is consistent with our hypotheses.

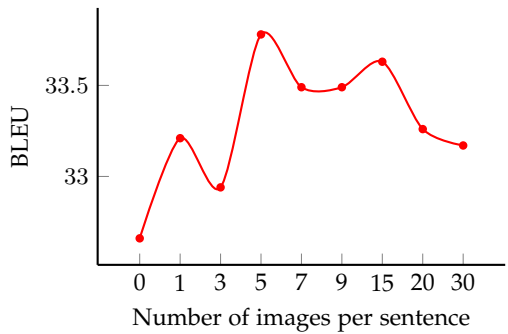
\begin{figure}[t]
\setlength{\abovecaptionskip}{0pt}
			\begin{center}
				
				\pgfplotsset{height=5cm,width=7cm,compat=1.15,every axis/.append style={thick},every axis legend/.append style={ at={(0.95,0.95)}},legend columns=1 row=2} 
				
				\begin{tikzpicture} \tikzset{every node}=[font=\small] 
				\begin{axis} [width=7cm,enlargelimits=0.13,legend pos=north west, xticklabels={0,1,3,5,7,9,15,20,30}, axis y line*=left,
				axis x line*=left, xtick={0,1,2,3,4,5,6,7,8}, 
				x tick label style={rotate=0},
				ylabel={BLEU},
				ylabel style={align=left},xlabel={Number of images per sentence},font=\small]

				\addplot+[smooth, mark=*,mark size=1.2pt,mark options={solid,mark color=red}, color=red] coordinates {(0, 32.66) (1, 33.21)  (2, 32.94)  (3, 33.78) (4, 33.49)  (5, 33.49)  (6, 33.63) (7, 33.26) (8, 33.17)};
				\end{axis}

				\end{tikzpicture}
			\end{center}
			\caption{\label{num_img}Influence of the number of images on the BLEU score.}
\end{figure}

\begin{table}
\centering
	\caption{Experiments on different source languages using the Multi30K dataset. The baseline is Transformer-Tiny.} 
	{
		\begin{tabular}{llllll}
			\toprule
             \multicolumn{1}{c}{\multirow{2}{*}{\textbf{Model}}} & \multicolumn{2}{c}{\textbf{De-En}} &   & \multicolumn{2}{c}{\textbf{Fr-En}}    \\ 
            \cmidrule{2-3} \cmidrule{5-6}
			\multicolumn{1}{c}{} & \textbf{Test2016} & \textbf{Test2017}  & &\textbf{Test2016} & \textbf{Test2017} \\ 
			\midrule
		     Baseline   &   42.88  &  40.57          &   &    54.60   &    48.45           \\
			 UVR-TILT  &  43.10     &  40.07    &   & 54.92     &  49.10      \\
			\bottomrule
		\end{tabular}
	}
\label{tbl:mmt_lang}
\end{table}

\subsection{Effectiveness across languages}
Table \ref{tbl:mmt_lang} shows the experiment results on different source languages. We observe that our method is applicable when the source texts are in other languages such as German and French. Our proposed methods are supposed to be independent of languages because the calculation for image retrieval only relies on the light lookup table, which can be extracted from an off-the-shelf seed corpus that is available for many languages.

\subsection{Joint Training and Fine-tuning}\label{sec:joint}

Since the multimodal and text-only machine translation tasks can benefit from the visual modality after retrieving images from the seed corpus, it is possible to bridge both tasks to train an even more powerful model. The connections between the texts and images inside the Multi30K datasets could be strong indications to bridge the gap between text and image modalities.

Therefore, we train a unified model based on UVR-CMRM by using the joint En-De datasets of Multi30K and WMT'14 (\textit{Joint Model}), and respectively train the Multi30K (\textit{Fine-tuned Multi30K}) and WMT'14 (\textit{Fine-tuned WMT}) models by initializing the trainable model parameters using the joint model. According to the results shown in Table \ref{tbl:jnt}, we summarize the following observations: 

(i) Two-stage training (joint training and fine-tuning) can boost the performance on the two concerned datasets, which indicates that the highly relevant Multi30K dataset can play the role of the seed data for training a text-only NMT model using our topic-image lookup table. 

(ii) The major gain is achieved by fine-tuning the smaller dataset, showing that the large-scale text-only dataset with our lookup table can also provide valuable complementary information for training a multimodal model on a much smaller dataset. The result further verifies the effectiveness of our method in low-resource settings.

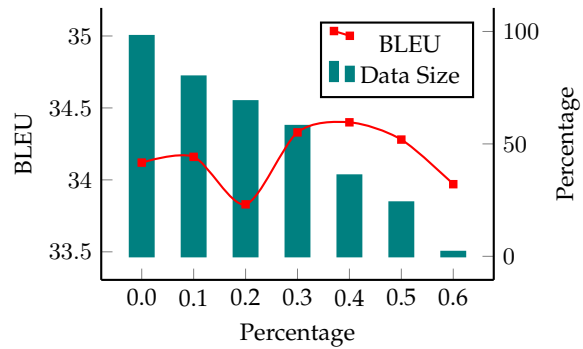
\begin{figure}[t]
\setlength{\abovecaptionskip}{0pt}
\begin{center}
\pgfplotsset{height=5.2cm,width=6.8cm,compat=1.14,every axis/.append style={thick},every axis legend/.append style={ at={(0.95,0.95)}},legend columns=1 row=2} 

\begin{tikzpicture} \tikzset{every node}=[font=\small]

\begin{axis}[hide x axis,
                 axis y line*=right,
                 axis x line*=right,
                 ybar=5pt,
                 width=6.8cm,
                 bar width=8pt,
                 enlargelimits=0.13,
                 ylabel={Percentage},
                 font=\small,
                 symbolic x coords={0,1,2,3,4,5,6},
                 xtick=data,
                 bar width=9pt,
                 ],
                 \addlegendimage{/pgfplots/refstyle=plot_one}\addlegendentry{BLEU}
                 \addplot+[teal] coordinates {(0,98) (1,80) (2,69) (3,58) (4,36) (5,24) (6,2)};
                 \addlegendentry{\small Data Size}
                 \end{axis}
                 
                 \begin{axis} [width=6.8cm,enlargelimits=0.13, xticklabels={0.0, 0.1, 0.2, 0.3, 0.4, 0.5, 0.6}, axis y line*=left, axis x line*=left, ymin=33.5, ymax=35, xtick={0,1,2,3,4,5,6}, x tick label style={rotate=0},
  ylabel={BLEU},
  ylabel style={align=left},xlabel={Percentage},font=\small]
\addplot+ [smooth, mark=square*,mark size=1.2pt,mark options={mark color=cyan}, color=red] coordinates { (0,34.12) (1,34.16) (2,33.83) (3,34.33) (4,34.40) (5,34.28) (6,33.97)};
\label{plot_one}
\end{axis}

\end{tikzpicture}
\end{center}
  \caption{BLEU score for different similarity thresholds.}
  \label{sim_thred}
\end{figure}

\begin{table}
\centering
	{
		\caption{Results of joint training and fine-tuning.}
		\label{tbl:jnt}} 
	{
	\begin{tabular}{lcc}
		\toprule
		\textbf{Model} &  \textbf{Multi30K Task}  & \textbf{WMT Task}  \\ 
		\midrule
        Joint training baseline & 37.28 & 27.68 \\
        \quad + Fine-tuned Multi30K & - & 27.96  \\
        \quad + Fine-tuned WMT & 43.13 & - \\
		\bottomrule
	\end{tabular}
	}

\end{table}

\subsection{Parameter Sensitivity Analysis}\label{sec:sensitivity}
In this section, we analyze our model sensitivity against parameter settings, including similarity threshold, number of images, and gating weight.

\textbf{Influence of the similarity threshold.}\label{sec:thred}
We investigate the influence of the similarity threshold $\delta$ that is set to filter the top-ranked images for each sentence in UVR-CMRM. Figure \ref{sim_thred} shows the performance for thresholds in [0.0, 0.1, 0.2, 0.3, 0.4, 0.5, 0.6] on the En-Ro test set. We observe that setting the threshold around 0.4 can yield a good balance of data size and BLEU score. It is reasonable that the best thresholds vary for different datasets because of the domain divergence of the image corpus for pre-training. 

\textbf{Influence of the number of images.}
To evaluate the influence of the number of paired images $m$ for UVR-TILT, we constrain $m$ in \{0, 1, 3, 5, 7, 9, 15, 20, 30\} for experiments on the En-Ro test set, as shown in Figure~\ref{num_img}.
When $m=0$, the model is the baseline NMT model, whose BLEU score is lower than all the models with images. 
As the number of images increases, the BLEU score also increases at the beginning (from $32.66$ to $33.78$) and then slightly decreases when $m$ exceeds 5. The reason might be that too many images for a sentence would have a higher chance of noise. Therefore, we set $m=5$ in our models. 

\begin{figure}
\setlength{\abovecaptionskip}{0pt}
			\begin{center}
				
				\pgfplotsset{height=5cm,width=7cm,compat=1.15,every axis/.append style={thick},every axis legend/.append style={at={(0.95,0.95)}},legend columns=3 row=1} \begin{tikzpicture} \tikzset{every node}=[font=\small] 
				\begin{axis} [width=7cm,enlargelimits=0.13,legend pos=north west,legend style={nodes={scale=0.68, transform shape},at={(0.03,1.08)},anchor=west},xticklabels={0.1,0.3,0.5,0.7,0.9}, axis y line*=left, axis x line*=left, xtick={0,1,2,3,4}, x tick label style={rotate=0},
				ylabel={BLEU},
				ylabel style={align=left},xlabel={Weight $\lambda$},font=\small]
				+\addplot+ [smooth, mark=square*,mark size=1.2pt,mark options={mark color=cyan}, color=red] coordinates { (0,32.96) (1,33.12) (2,33.06) (3,33.47)(4,33.02)};
				\addlegendentry{\small Manual weight}
				\addplot+[densely dotted, mark=none, color=black] coordinates {(0, 32.66)(1, 32.66)(2, 32.66)(3, 32.66)(4, 32.66)};
				\addlegendentry{\small Transformer-Base}
				\addplot+[sharp plot, mark=none, color=cyan] coordinates {(0, 33.78)(1, 33.78)(2, 33.78)(3, 33.78)(4, 33.78)};
				\addlegendentry{\small Ours}
				\end{axis}
				\end{tikzpicture}
				\caption{\label{gate}Quantitative study of the gating weight $\lambda$.}
				
			\end{center}
\end{figure}
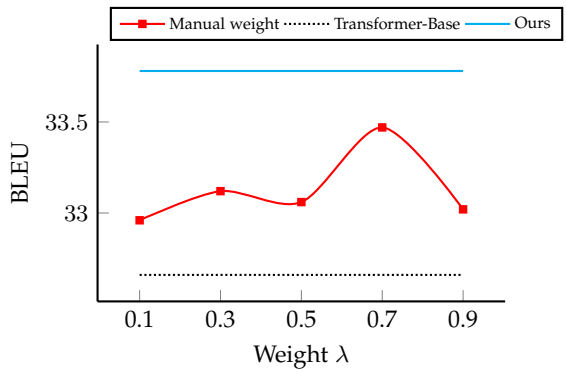

\textbf{Influence of gating weight $\lambda$.}\label{sec:gate}
In our model, the weight $\lambda$ of the gated aggregation method is learned automatically to measure the importance of the visual information. 
We compare by manually setting the weight $\lambda$ to scalar values in \{0.1, 0.3, 0.5, 0.7, 0.9\} for experiments of UVR-TILT on the En-Ro test set.
Figure~\ref{gate} shows that all models with manual $\lambda$ outperform the baseline Transformer-base, indicating the effectiveness of image information.
In contrast, they are inferior to the performance of our model.
This means that the degree of dependency for image information varies for each source sentence, indicating the necessity of automatically learning the gating weights of image representations.

\subsection{Computation Efficiency} \label{time}
There are mainly two extra computation costs using our method, including (i) obtaining image data for sentences and (ii) learning image representations, which are negligible compared with training an NMT model. The time of retrieving image data for MT sentences for the En-Ro dataset is less than 1 minute using GPU. The lookup table is formed as the mapping of the token (only topic words) index to the image id. Then, the retrieval method is applied as the tensor indexing from the sentence token indices (only topic words) to image ids, which is the same as the procedure of word embedding. The retrieved image ids are then sorted by frequency. Learning image representations takes about 2 minutes for all the 29,000 images in Multi30K using 6G GPU memory for feature extraction and eight CPU threads for transforming images. The extracted features are formed as the ``image embedding layer" in the size of (29000, 2400) for quick access in the neural network.

\section{Conclusions}
This work investigates a flexible framework to incorporate visual information into sentence modeling by image retrieval from a light lookup table and learned cross-modal embedding space. Extensive empirical experiments on 14 benchmark datasets verify the effectiveness of the proposed method. A series of case studies are conducted to evaluate visual benefits and influence factors. Our method is general and can be easily implemented in existing deep-learning NLP systems for different languages. Through the proposed retrieval methods, we can provide a group of images that disclose a diversity of implicit topics that might be entailed in sentences, yielding better context grounding with fine-grained information. We show that our method enriches the representation of content words, {provide fine-grained grounding information about the relationship between concepts and events}, and potentially enhances the accuracy of disambiguation. Besides incorporating images to build the pairwise relationship across modalities, it is potential to incorporate various extra knowledge as alignment topic information in the future, such as audio, not only images. 

\section*{Acknowledgement}
Part of this study has been published as ``Neural Machine Translation with Universal Visual Representation" \cite{Zhang2020Neural} in the Eighth International Conference on Learning Representations (ICLR 2020). The extension includes three sides: (i) general tasks: this work studies the universal visual representation for language representation in a broader view of the natural language processing scenario, with experiments on 14 representative NLP tasks; (ii) new method: this work investigates new methods of semantic sentence-image matching from a shared cross-modal space, to give more accurately paired images as topic information; (iii) in-depth analysis to interpret the benefits from the visual modality. 


\ifCLASSOPTIONcaptionsoff
  \newpage
\fi

\bibliography{uvm}
\bibliographystyle{IEEEtranN}

\vspace{-12mm}
	\begin{IEEEbiography}[{\includegraphics[width=1in,height=1.25in,clip,keepaspectratio]{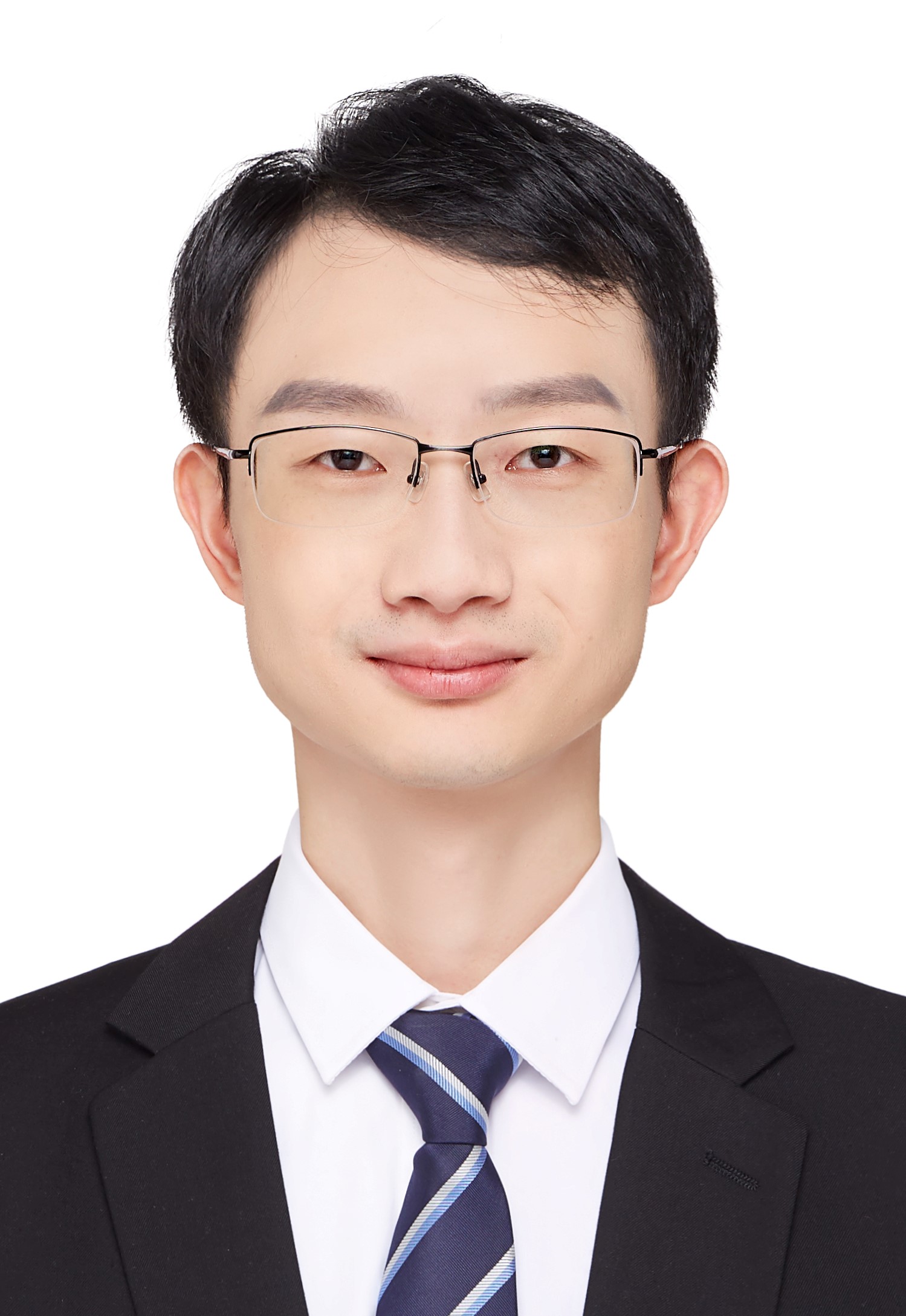}}]{Zhuosheng Zhang}
  		received his Bachelor's degree in internet of things from Wuhan University in 2016, his M.S. degree in computer science from Shanghai Jiao Tong University in 2020. He is working towards his Ph.D. degree in computer science with the Center for Brain-like Computing and Machine Intelligence of Shanghai Jiao Tong University. He was an internship research fellow at NICT from 2019-2020. His research interests include natural language processing, machine reading comprehension, dialogue systems, and machine translation. 
	\end{IEEEbiography}
\vspace{-8mm}
\begin{IEEEbiography}[{\includegraphics[width=1in,height=1.25in,clip,keepaspectratio]{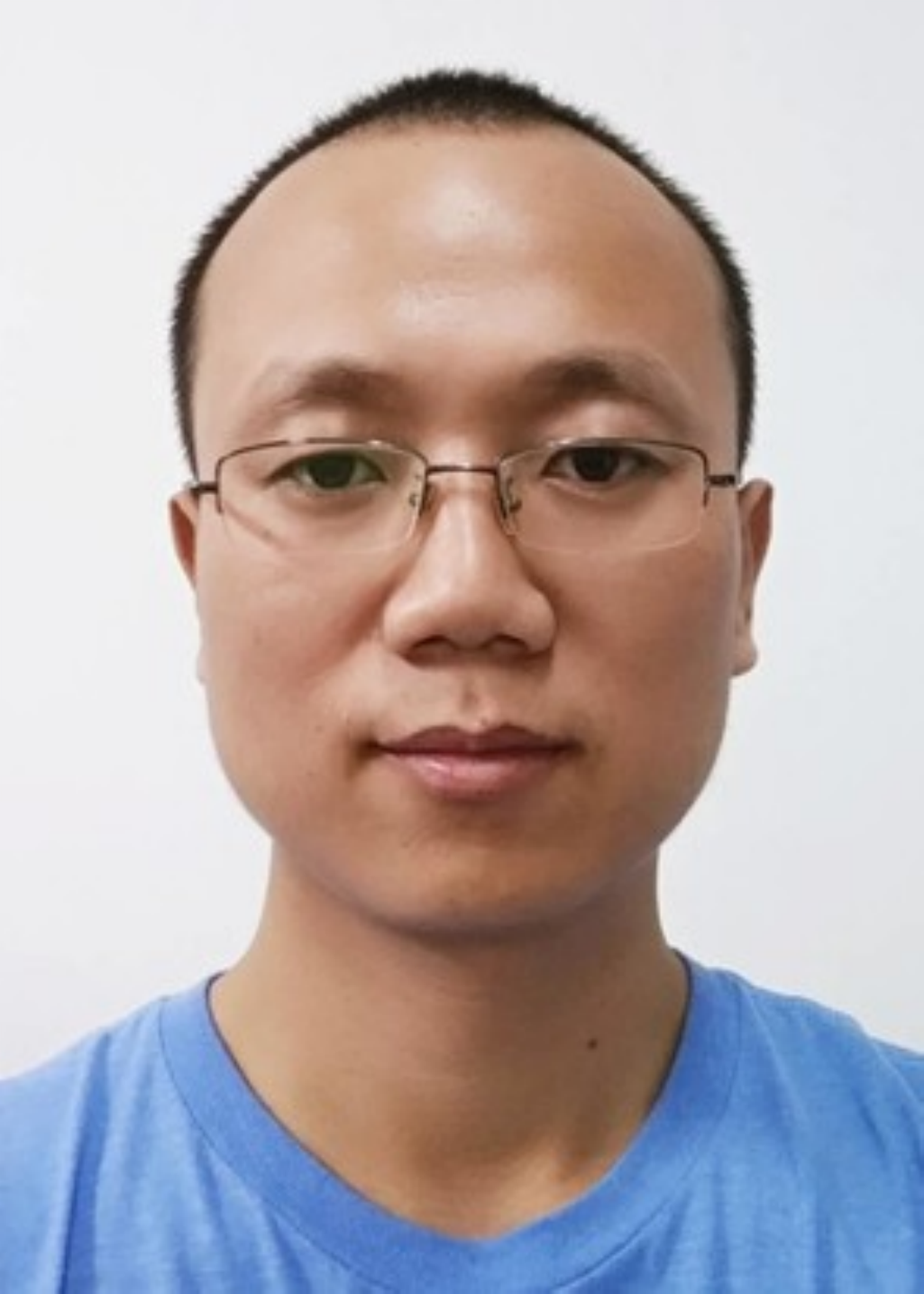}}]{Kehai Chen}
is an Assistant Professor at Harbin Institute of Technology (Shenzhen) since 2022. Before that, he was a researcher at Japan National Institute of Information and Communications Technology (NICT) from 2018 to 2021. He received the Ph.D. degree in computer science from Harbin Institute of Technology in 2018. His research interests include machine translation and natural language processing.
\end{IEEEbiography}
\vspace{-12mm}
\begin{IEEEbiography}[{\includegraphics[width=1in,height=1.25in,clip,keepaspectratio]{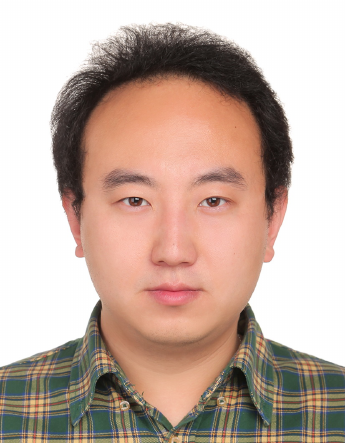}}]{Rui Wang}
	is an associate professor at Shanghai Jiao Tong University since 2021. Before that, he was a researcher (tenured in 2020) at Japan National Institute of Information and Communications Technology (NICT) from 2016 to 2020. He received his B.S. degree from Harbin Institute of Technology in 2009, his M.S. degree from the Chinese Academy of Sciences in 2012, and his Ph.D. degree from Shanghai Jiao Tong University in 2016, all of which are in computer science. He was a joint Ph.D. at Centre Nationnal de la Recherche Scientifique, France in 2014. His research interests are machine translation and natural language processing.
\end{IEEEbiography}

\vspace{-12mm}

\begin{IEEEbiography}[{\includegraphics[width=1in,height=1.25in,clip,keepaspectratio]{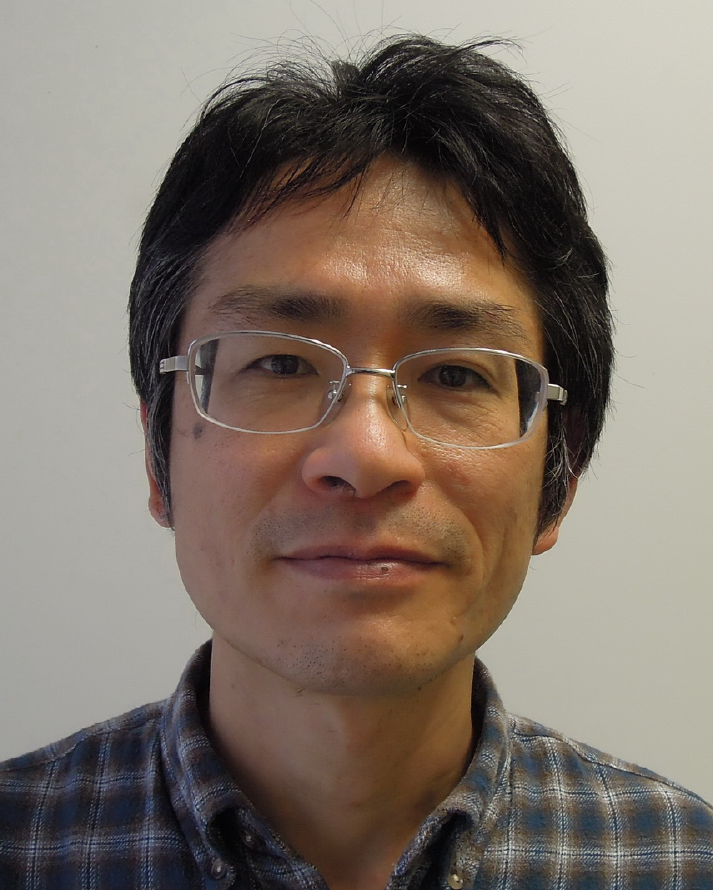}}]{Masao Utiyama} is a research manager of the National Institute of Information and Communications Technology, Japan. He completed his doctoral dissertation at the University of Tsukuba in 1997. His main research field is machine translation.
\end{IEEEbiography}

\vspace{-12mm}

\begin{IEEEbiography}[{\includegraphics[width=1in,height=1.25in,clip,keepaspectratio]{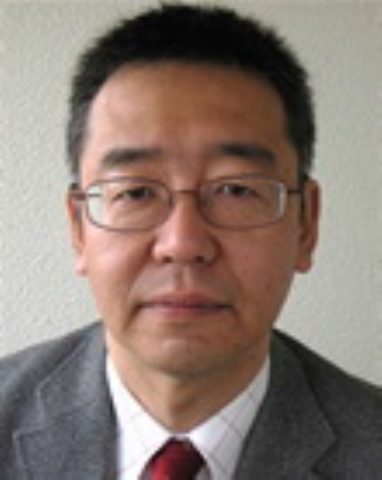}}]{Eiichiro Sumita}
received the Bachelor and Master degree in computer science from The University of Electro-Communications, Japan in 1980 and 1982 and the  Ph.D degree in Engineering from  Kyoto University, Japan in 1999. He is currently Director of Multilingual Translation Laboratory of National Institute of Information and Communication Technology from 2006. He worked at Advanced Telecomunications Research Institute International from 1992 to 2009 and IBM Research-Tokyo from 1980 to 1991. His research interests
include machine translation and e-Learning.
\end{IEEEbiography}

\vspace{-12mm}

\begin{IEEEbiography}[{\includegraphics[width=1in,height=1.25in,clip,keepaspectratio]{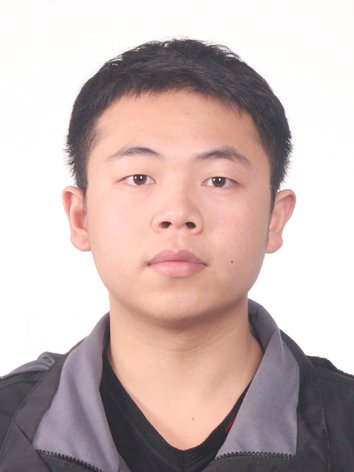}}]{Zuchao Li}
 received the B.S. degree from Wuhan University, Wuhan, China, in 2017. Since 2017, he has been a Ph.D. student with the Center for Brain-like Computing and Machine Intelligence of Shanghai Jiao Tong University, Shanghai, China. His research focuses on natural language processing, especially syntactic and semantic parsing. 
\end{IEEEbiography}

\vspace{-12mm}

\begin{IEEEbiography}[{\includegraphics[width=1in,height=1.25in,clip,keepaspectratio]{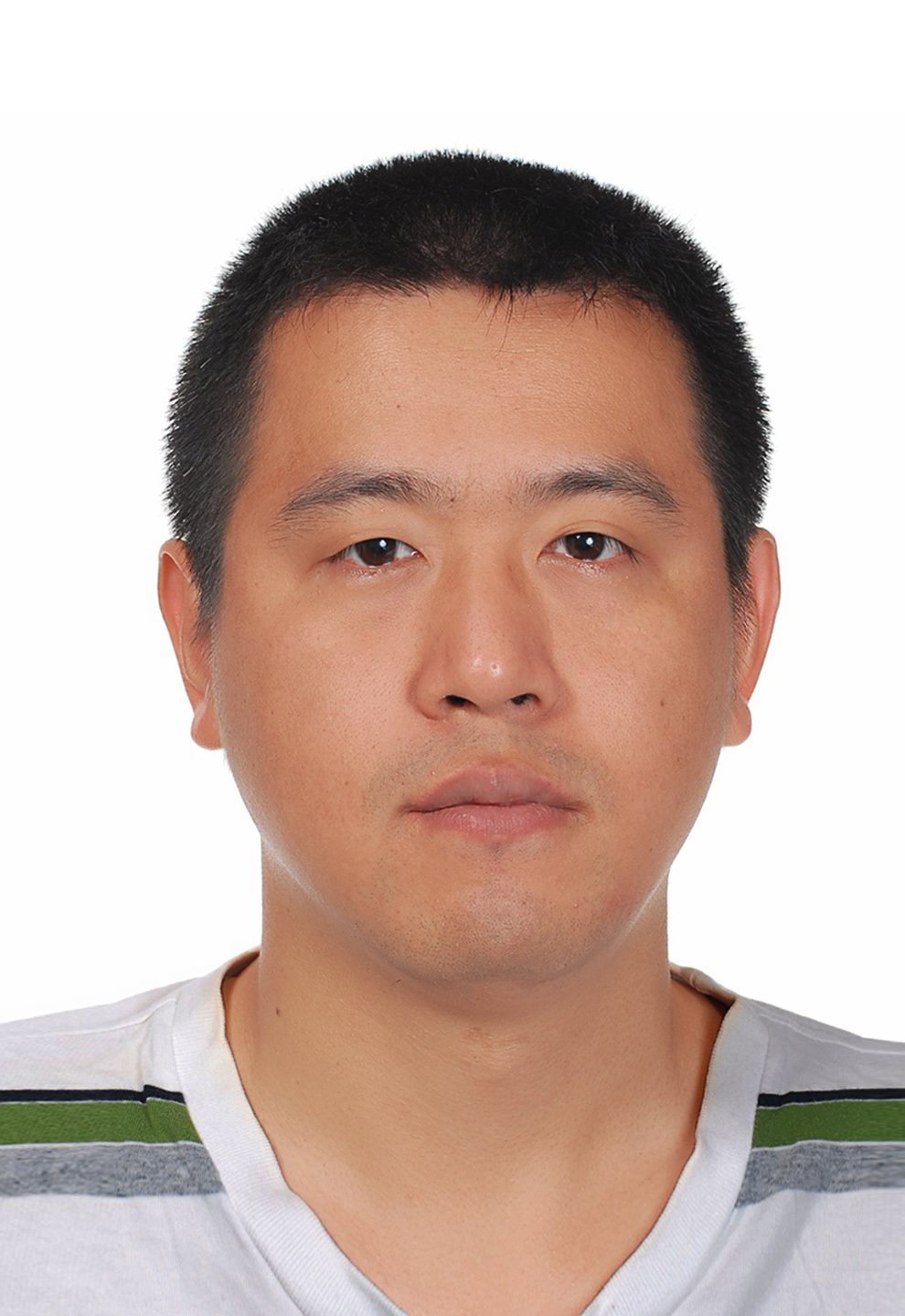}}]{Hai Zhao}
	received the BEng degree in sensor and instrument engineering, and the MPhil degree in control theory and engineering from Yanshan University in 1999 and 2000, respectively,
	and the PhD degree in computer science from Shanghai Jiao Tong University, China in 2005. 
	He is currently a full professor at department of computer science and engineering,  Shanghai Jiao Tong University after he joined the university in 2009. 
	He was a research fellow at the City University of Hong Kong from 2006 to 2009, a visiting scholar in Microsoft Research Asia in 2011, a visiting expert in NICT, Japan in 2012.
	He is an ACM professional member, and served as area co-chair in ACL 2017 on Tagging, Chunking, Syntax and Parsing, (senior) area chairs in ACL 2018, 2019 on Phonology, Morphology and Word Segmentation.
	His research interests include natural language processing and related machine learning, data mining and artificial intelligence.
\end{IEEEbiography}
\end{CJK*}
\end{document}